\documentclass[lettersize,journal]{IEEEtran}

\usepackage{amsmath,amssymb,amsfonts}
\usepackage{algorithmic}
\usepackage{algorithm}
\usepackage{array}
\usepackage{cite}
\usepackage{graphicx}
\usepackage{xcolor}
\usepackage[hyphens]{url}
\usepackage[hidelinks,pdftex]{hyperref}
\usepackage[all]{hypcap}
\usepackage[T1]{fontenc}
\usepackage{multirow}
\usepackage{stfloats}
\usepackage{textcomp}
\usepackage{verbatim}
\usepackage{booktabs}
\usepackage{subcaption}
\usepackage{wrapfig}
\usepackage{enumitem}
\usepackage{balance}

\def\BibTeX{{\rm B\kern-.05em{\sc i\kern-.025em b}\kern-.08em
    T\kern-.1667em\lower.7ex\hbox{E}\kern-.125emX}}

\begin{document}
\hypersetup{
	pdfinfo ={
			Title={Space Robotics Bench: Robot Learning Beyond Earth},
			Author={Andrej Orsula, Matthieu Geist, Miguel Olivares-Mendez, Carol Martinez},
			Keywords={Space Robotics, Simulation, Robot Learning},
			Creator={LaTeX},
		}
}

\title{%
    \fontsize{23.99999}{25.8}\selectfont%
    Space Robotics Bench: Robot Learning Beyond Earth%
}

\author{
	Andrej Orsula\textsuperscript{1},
	Matthieu Geist\textsuperscript{2},
	Miguel Olivares-Mendez\textsuperscript{1},
	Carol Martinez\textsuperscript{1}%
	\thanks{\textsuperscript{1}University of Luxembourg}%
	\thanks{\textsuperscript{2}Earth Species Project}%
}

\maketitle

\begin{abstract}
  The growing ambition for space exploration demands robust autonomous systems that can operate in unstructured environments under extreme extraterrestrial conditions. The adoption of robot learning in this domain is severely hindered by the prohibitive cost of technology demonstrations and the limited availability of data. To bridge this gap, we introduce the Space Robotics Bench, an open-source simulation framework for robot learning in space. It offers a modular architecture that integrates on-demand procedural generation with massively parallel simulation environments to support the creation of vast and diverse training distributions for learning-based agents. To ground research and enable direct comparison, the framework includes a comprehensive suite of benchmark tasks that span a wide range of mission-relevant scenarios. We establish performance baselines using standard reinforcement learning algorithms and present a series of experimental case studies that investigate key challenges in generalization, end-to-end learning, adaptive control, and sim-to-real transfer. Our results reveal insights into the limitations of current methods and demonstrate the utility of the framework in producing policies capable of real-world operation. These contributions establish the Space Robotics Bench as a valuable resource for developing, benchmarking, and deploying the robust autonomous systems required for the final frontier.
\textit{The source code is available at~\href{https://github.com/AndrejOrsula/space_robotics_bench}{github.com/AndrejOrsula/space\_robotics\_bench}.}

\end{abstract}

\begin{IEEEkeywords}
  Space Robotics, Simulation, Robot Learning

\end{IEEEkeywords}

\section{Introduction}
\label{sec:introduction}

\IEEEPARstart{R}{obots} are poised to play a transformative role in the future of space exploration and utilization. Ambitious missions envision fleets of autonomous rovers exploring planetary surfaces~\cite{jpl2024enabling} and robotic manipulators constructing large orbital structures~\cite{rognant2019autonomous, zhihui2021review}. The success of these long-duration endeavors depends on developing robotic systems that can operate reliably and adapt to unstructured conditions with minimal human supervision. Robot learning, particularly reinforcement learning~(RL)~\cite{sutton2018reinforcement}, offers a powerful paradigm for acquiring such adaptive behaviors~\cite{gu2025humanoid, kim2024openvla}. However, the adoption of these data-intensive methods in space robotics is severely hindered by significant challenges. Extraterrestrial environments suffer from extreme data scarcity, while real-world technology demonstrations are prohibitively expensive. Furthermore, existing space-grade simulators are often narrow in scope or remain inaccessible to the broader research community, creating a substantial barrier to innovation.

A fundamental challenge for any simulation-based approach is the sim-to-real gap. A common strategy is to develop a singular, high-fidelity digital twin of a target environment. This approach is effective for verification in well-defined scenarios but becomes inherently risky for unknown and unpredictable extraterrestrial conditions. A policy trained in such a singular environment may overfit to unverified assumptions about the world. This can result in a brittle system that fails when faced with real-world variations. We propose an alternative paradigm that shifts the focus from the pursuit of perfect simulation fidelity to the more robust goal of mastering a massive diversity of scenarios. By exposing an agent to a vast range of environments created through procedural content generation~(PCG)~\cite{cobbe2020leveraging} and domain randomization~(DR)~\cite{tobin2017domain}, it is forced to learn a generalizable understanding of the underlying physics. In turn, this approach treats the sim-to-real gap as a distribution of potential realities to be encountered and overcome.

\begin{figure}[t]
    \vspace{0.225em}
    \centering
    \includegraphics[width=\linewidth]{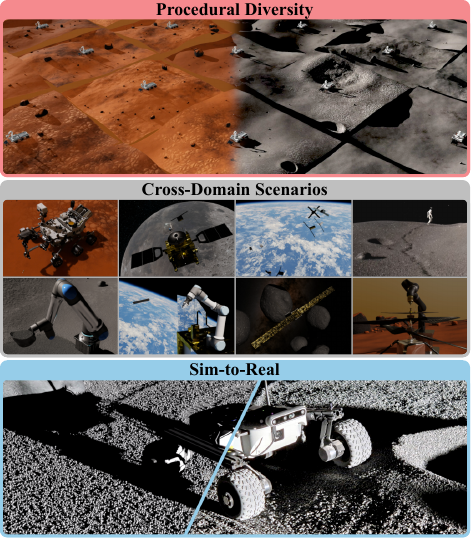}
    \caption{Space Robotics Bench brings robot learning to space by leveraging procedural diversity across a wide range of cross-domain scenarios. The entire workflow is validated through a successful zero-shot sim-to-real transfer to a physical robot.}
    \label{fig:overview}
    \vspace{-1.0em}
\end{figure}

To realize this paradigm, we introduce the Space Robotics Bench~(SRB), an open-source simulation framework designed from the ground up to accelerate research in robotic autonomy for space applications. As illustrated in Fig.~\ref{fig:overview}, SRB supports the complete end-to-end workflow of our proposed approach. It facilitates a variety of development workflows, from training RL agents to collecting demonstration data for imitation learning and validating traditional control systems. Its core strength lies in its ability to generate a virtually unlimited number of unique training scenarios. This is accomplished through extensive PCG for scenes and embodiments, coupled with comprehensive randomization of physical and visual parameters. To make training on this diverse data feasible, SRB leverages massive parallelism enabled by its GPU-accelerated backend. Moreover, the entire framework is built upon standard community APIs to ensure accessibility and ease of use.

This work presents SRB as a complete and validated workflow for developing robust autonomous systems. We introduce a modular framework tailored for robot learning in space. We present a suite of benchmark tasks designed to evaluate autonomous capabilities across a wide range of mission-relevant scenarios. We also establish baselines that compare the performance of standard RL algorithms to ground future work. The central contribution is the empirical validation of our procedural paradigm. We demonstrate its effectiveness through a series of case studies, culminating in the successful zero-shot sim-to-real transfer of a learned navigation policy. Through these contributions, we aim to provide a valuable resource for developing the robust autonomous systems required for the final frontier.

\section{Related Work}
\label{sec:related_work}

This work is positioned at the intersection of space simulation, robot learning benchmarks, and techniques for generalization. The contribution of SRB is its unique synthesis of these domains to address a critical gap in the development of robust, learning-based autonomy for space.

\subsection{Space Simulation Frameworks}

\begin{table*}[b]
    \centering
    \caption{\textsc{Comparison of SRB with Existing Space Simulation Frameworks.}}
    \label{tab:simulator_comparison}
    \begin{tabular}{@{}lllccccc@{}}
        \toprule
        \textbf{Category}                        & \textbf{Simulator}                        & \textbf{Primary Focus}         & \textbf{Accessibility} & \textbf{Extensibility} & \textbf{Sim-to-Real} & \textbf{Scalability} & \textbf{Gymnasium}                            \\
        \midrule
        \multirow{2}{*}{\textit{Astrodynamics}}  & GMAT~\cite{hughes2014verification}        & Orbital Mechanics              & Open                   & Low                    & N/A                  & Low                  & No                                            \\
                                                 & Basilisk~\cite{kenneally2020basilisk}     & Orbital Mechanics              & Open                   & Medium                 & N/A                  & Medium               & \hspace{5.75mm}Yes~\cite{stephenson2024bskrl} \\
        \midrule
        \multirow{5}{*}{\textit{Mission V\&V}}   & VIPER RSIM~\cite{stukes2021innovative}    & Rover V\&V                     & Restricted             & Low                    & Digital Twin         & Low                  & No                                            \\
                                                 & HeliCAT-DARTS~\cite{largescale2024heli}   & Rotorcraft V\&V                & Restricted             & Low                    & Digital Twin         & Low                  & No                                            \\
                                                 & EELS-DARTS~\cite{hardeers2024eelsdarts}   & Snake Robot V\&V               & Restricted             & Low                    & Digital Twin         & Low                  & No                                            \\
                                                 & Astrobee Sim~\cite{coltin2019astrobee}    & Free-Flyer V\&V                & Open                   & Medium                 & Digital Twin         & Low                  & No                                            \\
                                                 & Int-Ball2 Sim~\cite{Hirano2025IntBall2}   & Free-Flyer V\&V                & Open                   & Low                    & Digital Twin         & Low                  & No                                            \\
        \midrule
        \multirow{6}{*}{\textit{Robot Learning}} & OmniLRS~\cite{antoine2024omnilrs}         & Planetary Navigation           & Open                   & Medium                 & Photorealism         & Medium               & No                                            \\
                                                 & RLRoverLab~\cite{mortensen2024rlroverlab} & Planetary Navigation           & Open                   & Low                    & DR                   & High                 & Yes                                           \\
                                                 & RANS~\cite{el2023rans}                    & Spacecraft Navigation          & Open                   & Low                    & DR                   & High                 & Yes                                           \\
                                                 & SpaceRobotEnv~\cite{wang2022collision}    & Orbital Manipulation           & Open                   & Low                    & DR                   & Low                  & Yes                                           \\
                                                 & GraspPlanetary~\cite{orsula2022learning}  & Planetary Manipulation         & Open                   & Low                    & DR                   & Low                  & Yes                                           \\
                                                 & \textbf{SRB (our)}                        & \textbf{Multi-Domain Autonomy} & \textbf{Open}          & \textbf{High}          & \textbf{DR+PCG}      & \textbf{High}        & \textbf{Yes}                                  \\
        \bottomrule
    \end{tabular}
\end{table*}

Simulation is a cornerstone of space mission development, traditionally used for high-fidelity verification of astrodynamics via tools like GMAT~\cite{hughes2014verification} and Basilisk~\cite{kenneally2020basilisk, stephenson2024bskrl}. However, space robotics introduces challenges that require physically realistic models of contact dynamics, complex terrains, and sensor feedback. To meet these needs, a new generation of specialized simulators has emerged, often developed by space agencies for specific, high-stakes missions. For example, the VIPER mission utilizes a dedicated Rover Simulation Software~(RSIM) for its software life cycle development~\cite{stukes2021innovative}. Similarly, HeliCAT-DARTS was developed for the life cycle of the Ingenuity Mars Helicopter~\cite{largescale2024heli}, and EELS-DARTS supports the development of snake-like robots for planetary exploration~\cite{hardeers2024eelsdarts}. Free-flying robots operating inside the International Space Station~(ISS), such as Astrobee~\cite{coltin2019astrobee} and Int-Ball2~\cite{Hirano2025IntBall2}, also rely on custom simulators for ground testing and verification.

These mission-specific simulators are invaluable for their intended purpose of verification and validation~(V\&V) applied to a single robotic system in a well-defined target environment. However, this singular focus makes them ill-suited for the broader needs of robot learning research. They are designed to model a static, singular reality with maximum achievable fidelity, which is not well aligned with the massive data generation and environmental diversity required by modern robot learning algorithms. The necessary hardware validation for these simulators is often infeasible for academic researchers and is typically undertaken by large space agencies, in which case the corresponding simulators often maintain restricted access. In a technological field that is already conservative, these restrictions further hinder the advancements in robust autonomy.

In response, the research community has developed several learning-focused simulators. Frameworks such as OmniLRS~\cite{antoine2024omnilrs} prioritize photorealistic rendering for a single target environment, a crucial step for training perception systems. Other valuable platforms provide tools for learning control policies in specific operational domains, including navigation for rovers~\cite{mortensen2024rlroverlab} and spacecraft~\cite{el2023rans}, or manipulation in orbit~\cite{wang2022collision} and on planetary surfaces~\cite{orsula2022learning}. While these platforms are crucial steps forward, they are often tightly coupled to the original research objectives and a single robotic platform. This makes it difficult to study generalization across different robots or applications. Consequently, these simulators are not architected to support the diverse range of robots and tasks needed to develop general-purpose policies. The focus on narrow problem domains, combined with the difficulty of physical validation, comes at the expense of the broad applicability and deep environmental diversity essential for training truly generalizable policies. Table~\ref{tab:simulator_comparison} summarizes this diverse landscape and highlights the distinct position of SRB.

\subsection{Terrestrial Robot Learning Benchmarks}

The field of robot learning has advanced significantly through the development of standardized benchmarks. These platforms have been invaluable for driving progress, allowing for the direct comparison of algorithms on common tasks. Initial efforts focused on general manipulation suites such as RLBench~\cite{james2020rlbench}, RoboHive~\cite{kumar2023robohive}, and Robosuite~\cite{zhu2020robosuite}, which standardized a range of tabletop interaction skills. More recent, specialized benchmarks have pushed the boundaries of complex skills, targeting long-horizon assembly with FurnitureBench~\cite{heo2023furniturebench}, dexterous manipulation with RoboPianist~\cite{zakka2023robopianist}, and whole-body locomotion with HumanoidBench~\cite{sferrazza2024humanoidbench}.

While this extensive body of work is foundational, it exhibits a profound terrestrial bias. The environments and tasks are confined to structured Earth-centric scenarios that do not capture the unique challenges of space. More importantly for the broader robot learning community, many of these benchmarks are maturing around a relatively narrow set of problem domains, primarily focused on tabletop manipulation. SRB complements these existing tools by introducing a fresh suite of challenges rooted in scenarios from the application domain of space that are fundamentally different from typical terrestrial tasks. It offers problems with varying levels of difficulty, from robust locomotion over deformable terrains to precision manipulation under orbital dynamics. This provides a new testbed for the robot learning community to evaluate the limits of current algorithms on tasks requiring advanced physical reasoning and generalization, moving beyond the well-explored domain of structured interaction.

\subsection{Procedural Generation and Domain Randomization}

The most significant hurdle for any simulation-based approach is the sim-to-real gap. A proven strategy for creating transferable policies is DR, which forces robustness by varying physical and visual parameters of the simulation during training~\cite{tobin2017domain}. We adopt this philosophy and advance it with PCG, a technique for algorithmically creating a near-infinite distribution of unique environment entities. By preventing an agent from overfitting to any single scenario, PCG promotes the learning of truly generalizable strategies.

While this strategy has proven effective for benchmarking generalization in simplified 2D environments~\cite{cobbe2020leveraging, koutras2021marsexplorer}, its application in complex 3D robotics is an active area of research. Recent terrestrial benchmarks have begun to incorporate procedural assets, for instance, to increase the diversity of objects from a static collection~\cite{han2024fetchbench}. A core contribution of SRB is to advance this paradigm at a fundamentally deeper and larger scale. Rather than relying on a fixed dataset of pre-generated assets, SRB employs an on-demand runtime pipeline that programmatically generates a unique and complex 3D world for each parallel simulation instance. This methodology provides a massive and diverse stream of data ideal for training high-capacity models. Furthermore, SRB uniquely extends the scope of PCG beyond environmental features like terrains and objects to the morphology of the robot itself. This includes procedural assets for controllable spacecraft and robot end-effectors. This presents a novel challenge for the robot learning community by providing a testbed for developing algorithms that can generalize across variations in both the world and their own embodiment, a necessity for adaptability during long-term space missions where hardware degradation is inevitable.

\vspace{1.0em}

In summary, SRB fills a distinct and unmet need. It complements mission-specific simulators by providing an accessible, open-source platform designed for the massive scale and diversity required by modern robot learning. It addresses the limitations of terrestrial benchmarks by incorporating space-relevant physics and tasks that introduce novel challenges for the community. Finally, it uniquely synthesizes PCG and DR as first-class principles into a comprehensive 3D framework. This synthesis enables a new generation of research focused on creating the generalizable policies required for the future of space exploration.

\section{Space Robotics Bench}
\label{sec:srb}

\begin{figure*}[t]
    \centering
    \includegraphics[width=\linewidth]{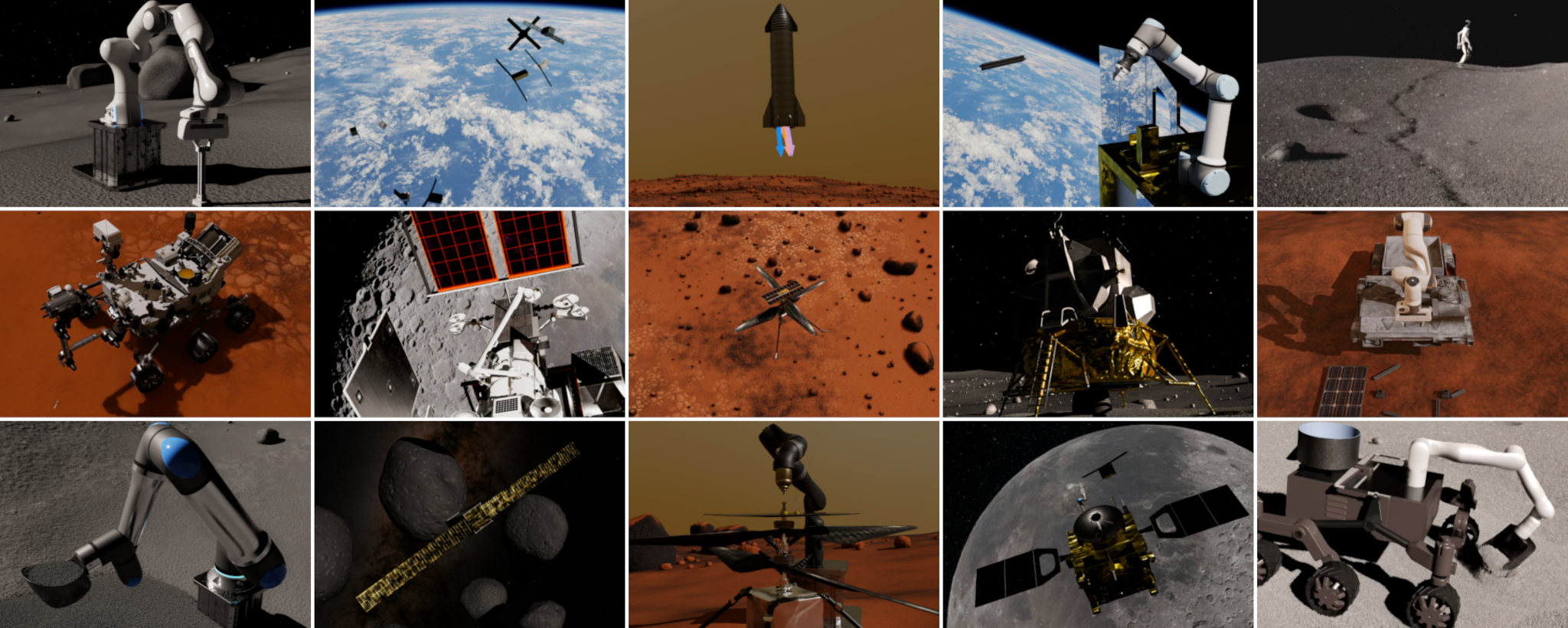}
    \caption{SRB supports diverse scenarios that encompass a wide range of challenges across multiple domains of space robotics.}
    \label{fig:showcase}
    \vspace{-1.0em}
\end{figure*}

SRB is a framework that provides both a versatile sandbox for research and a collection of benchmark tasks for space robotics. This section describes its core design as an open-source simulation framework built to accelerate the development of autonomy beyond Earth. The scenario diversity across multiple domains of space robotics is illustrated in Fig.~\ref{fig:showcase}.

\subsection{Framework Architecture}

SRB is built upon NVIDIA Isaac Sim and the Isaac Lab~\cite{mittal2023orbit} robot learning framework to leverage its powerful capabilities for parallelized physics and rendering. While Isaac Lab provides the foundation, SRB introduces a specialized architecture designed to realize the procedural paradigm for the unique demands of space robotics. SRB implements its own modular systems for asset management, scenario handling, and task logic that are tailored for extensibility. The design of the framework is guided by four key principles.

\paragraph{Modularity and Extensibility} SRB employs a modular architecture where every asset, robot, actuation model, and task is a self-contained component registered within an internal registry. This design allows researchers to easily introduce new elements by providing a standard model format and a data-validated configuration. This principle extends to the software itself, as the framework includes a streamlined sim-to-real mechanism that automates the translation of simulated task definitions into compact wrappers with modular hardware interfaces for deployment onto physical robots.

\paragraph{Accessibility and Integration} The framework is designed for broad accessibility and integration with established workflows. It provides a native interface for the ROS~2 middleware~\cite{macenski2022ros2}, and all tasks are compatible with the Gymnasium API~\cite{towers2024gymnasium}. This ensures seamless use with existing RL libraries such as Stable Baselines3~\cite{raffin2021stable} and skrl~\cite{serrano2023skrl}. A unified command-line interface offers users code-free control over most experimental parameters, including the selection of the task, domain, scenery, robot, action space, and the number of unique parallel environments.

\paragraph{Scale and Performance} SRB is engineered for high-throughput data acquisition, which is essential for modern learning algorithms. The GPU-accelerated capabilities of the underlying engine are enhanced with additional backend optimizations. These include Rust extension modules for high-performance CPU-bound logic via PyO3~\cite{pyo3}, and computationally expensive task logic is offloaded to the GPU using TorchScript~\cite{ansel2024pytorch} to maximize data throughput.

\paragraph{Fidelity and Diversity} Finally, SRB emphasizes the generation of diverse and realistic simulation environments. The framework combines explicit physics modeling with a strategy of enforced robustness. It incorporates space-relevant physics, including rigid body dynamics, particle interactions for granular materials, and variable gravity. For complex physical phenomena that are not explicitly modeled, such as atmospheric drag or turbulence, the framework approximates their impact by applying random disturbances in the form of external forces and torques, which forces an agent to learn a policy that is inherently robust to a wide range of unmodeled dynamics. This principle, combined with the on-demand generation of unique procedural assets and extensive DR, creates the rich training distribution detailed in the following subsections.

\subsection{Procedural Paradigm for Generalization}

\begin{figure}[!b]
    \vspace{-1.0em}
    \centering
    \includegraphics[width=\linewidth]{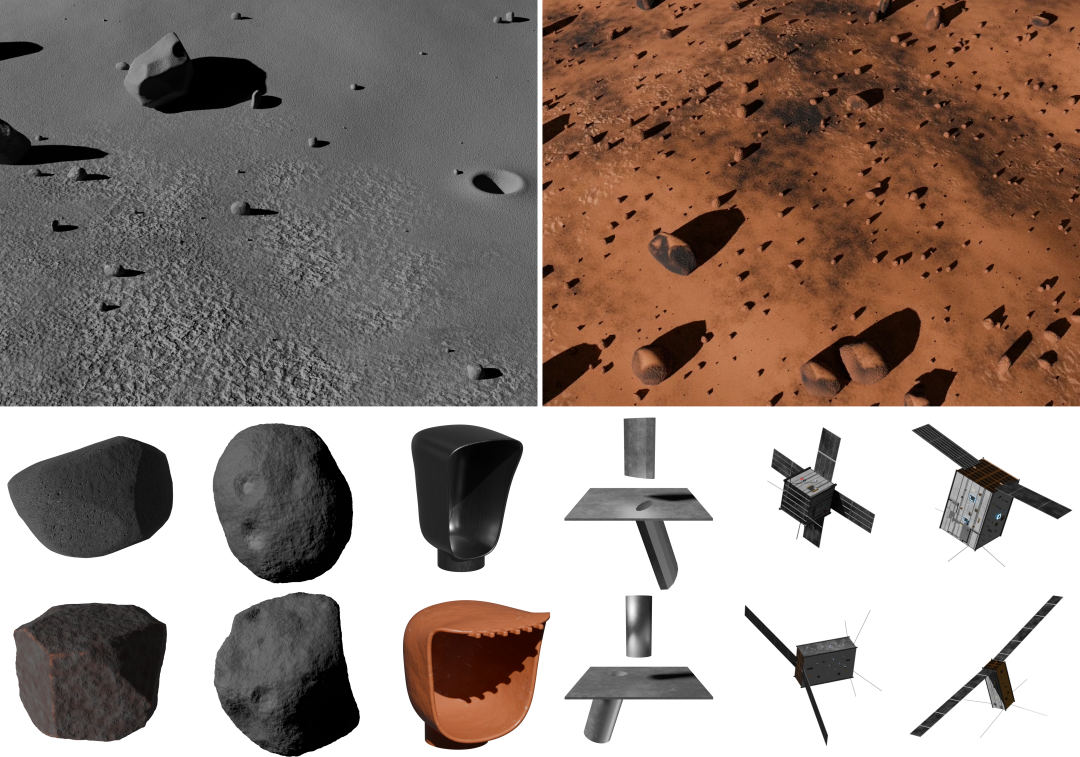}
    \caption{Examples of diverse PCG assets employed within SRB.}
    \label{fig:procgen_examples}
    \vspace{-0.375em}
\end{figure}

The strategy of SRB for generalization is built on two complementary layers of diversity. The first layer is PCG, which provides foundational structural novelty. To achieve this, SRB integrates a dedicated procedural engine that functions as an on-demand asset factory. It programmatically synthesizes a virtually unlimited variety of 3D assets using Blender~\cite{blender} as its generation backend. This engine leverages node-based parametric blueprints that define entire families of 3D models, and enable the creation of highly customizable environments. For example, the generation of a procedural lunar surface is a multi-stage process that builds complexity through layered proceduralism. The pipeline begins with a flat base mesh, which is first displaced by low-frequency Perlin noise to establish macro-scale topography. Subsequent layers of higher-frequency noise are then applied to carve out meso-scale features, such as using the cellular patterns of Voronoi noise to form the sharp rims of impact craters. Critically, this versatility extends beyond environmental features to the morphology of the robot itself. SRB includes procedural pipelines for mission-critical hardware, including satellites with randomized dimensions and excavation tools with unique geometries, as showcased in Fig.~\ref{fig:procgen_examples}.

This content is orchestrated through a seamless on-demand pipeline that programmatically links SRB to the procedural engine at runtime. Upon starting a simulation, SRB dispatches a request specifying the asset blueprints and unique seeds for each parallel environment. The engine then generates each unique asset variant in a headless background process. A fully automated export workflow programmatically bakes the procedural materials into standard Physically Based Rendering~(PBR) textures and exports the final asset in Universal Scene Description~(USD) format for efficient loading. The entire process enables the on-demand generation of hundreds of unique procedural worlds in a matter of seconds.

The second layer of diversity is extensive DR, which provides parametric novelty. Complementary to the structural diversity from PCG, SRB applies a comprehensive layer of DR at the start of each training episode. This technique systematically varies a wide range of physical and visual parameters to enforce robustness against unmodeled effects and sensor variations. Randomized physical parameters include gravity, object inertia, surface friction, and the cohesion of granular media. Visual parameters such as lighting conditions and sensor noise profiles are also varied. This two-layer approach of combining PCG with DR creates a massively parallel and highly diverse training distribution. It systematically prevents overfitting and forces an agent to learn a physically grounded and generalizable understanding of its task.

\subsection{Domains}

To ground its scenarios in realistic contexts, SRB introduces several template domains relevant to space activities, each with unique physical characteristics. In addition to the indicated gravity magnitude, the selected domain affects simulation aspects such as solar illumination, including the intensity, color temperature, and angular diameter of the Sun. While most of the SRB tasks are domain-agnostic to support cross-domain research, the framework provides the following presets:
\begin{list}{}{\leftmargin=0em}
    \item \textbf{Orbit:} Microgravity space environments\hfill\texttt{[g=0.00 m/s\textsuperscript{2}]}
    \item \textbf{Asteroid:} Small irregular bodies\hfill\texttt{[g=0.14\textpm0.14 m/s\textsuperscript{2}]}
    \item \textbf{Moon:} Lunar surface with regolith\hfill\texttt{[g=1.62\textpm0.01 m/s\textsuperscript{2}]}
    \item \textbf{Mars:} Martian surface with rocks\hfill\texttt{[g=3.72\textpm0.01 m/s\textsuperscript{2}]}
    \item \textbf{Earth:} Base terrestrial conditions\hfill\texttt{[g=9.81\textpm0.03 m/s\textsuperscript{2}]}
\end{list}

\subsection{Robots}

To populate these domains, SRB features a diverse and extensible collection of pre-configured robots. New robots can be added by providing a model in the USD format and a declarative Python class. This registers the robot with the asset management system and assigns it to a specific category. The inclusion of commercially available robots, while not space-rated, provides standardized platforms that are widely available and provide valuable insights for algorithm development and generalization studies. SRB collection currently includes:
\begin{list}{}{\leftmargin=0em\itemsep=0.25em}
    \item \textbf{Mobile Robot}
          \vspace{-0.25em}
          \begin{list}{}{\leftmargin=1.0em\itemsep=-0.1em}
              \item \textbf{Wheeled:} Planetary rovers (Perseverance, Pragyan)\hfill\texttt{| 4}
              \item \textbf{Legged:} Bipeds \& quadrupeds (Spot, ANYmal)\hfill\texttt{| 7}
              \item \textbf{Aerial:} Atmospheric rotorcraft (Ingenuity, Crazyflie)\hfill\texttt{| 2}
              \item \textbf{Spacecraft:} Satellites \& landers (ISS, PCG cubesat)\hfill\texttt{|12}
          \end{list}
    \item \textbf{Manipulator}
          \vspace{-0.25em}
          \begin{list}{}{\leftmargin=1.0em\itemsep=-0.1em}
              \item \textbf{Serial:} Robot arms (Canadarm, Franka, Kinova, UR)\hfill\texttt{|14}
          \end{list}
    \item \textbf{End-Effector} --- Attached to any \textbf{manipulator}
          \vspace{-0.25em}
          \begin{list}{}{\leftmargin=1.0em\itemsep=-0.1em}
              \item \textbf{Active:} Actuated tools (gripper, dexterous hand)\hfill\texttt{| 8}
              \item \textbf{Passive:} Non-actuated tools (screwdriver, scoop)\hfill\texttt{| 9}
          \end{list}
    \item \textbf{Mobile Manipulator}
          \vspace{-0.25em}
          \begin{list}{}{\leftmargin=1.0em\itemsep=-0.1em}
              \item \textbf{Humanoid:} Anthropomorphic robots (Unitree H1/G1)\hfill\texttt{| 4}
              \item \textbf{Combined:} Any \textbf{mobile robot + manipulator}\hfill\texttt{| \texttimes}
          \end{list}
\end{list}
A key feature of this system is its compositional design, which allows researchers to dynamically create novel mobile manipulators by combining any mobile base, manipulator arm, and end-effector. While the framework technically supports the programmatic creation of over 5000 unique mobile manipulators, such as mounting a large manipulator with a dexterous hand on a micro-drone, not all configurations are physically sensible. Yet, this flexibility is a deliberate design choice as it empowers researchers to explore a vast design space and systematically study how different hardware configurations impact task performance and policy generalization.

\subsection{Actuation Models}

SRB provides a library of modular actuation models that map normalized agent actions to low-level robot commands. A core feature is the automatic composition of these models for multi-component robots. An agent controlling a mobile manipulator with an active gripper will interface with a unified action space that combines the individual models for the base, arm, and end-effector. This design simplifies the control of complex systems. While direct joint-space control is supported, the framework emphasizes high-level, robot-agnostic abstractions to improve generalization. These include:
\begin{list}{}{\leftmargin=0em\itemsep=0.25em}
    \item \textbf{Wheeled}
          \vspace{-0.25em}
          \begin{list}{}{\leftmargin=1.0em\itemsep=-0.1em}
              \item \textbf{a)} Target linear \& angular velocity mapped via Kinematics
              \item \textbf{b)} Target joint velocities
          \end{list}
    \item \textbf{Legged \& Humanoid}
          \vspace{-0.25em}
          \begin{list}{}{\leftmargin=1.0em\itemsep=-0.1em}
              \item \textbf{a)} Target joint positions
          \end{list}
    \item \textbf{Aerial}
          \vspace{-0.25em}
          \begin{list}{}{\leftmargin=1.0em\itemsep=-0.1em}
              \item \textbf{a)} Target linear \& angular accelerations
          \end{list}
    \item \textbf{Spacecraft}
          \vspace{-0.25em}
          \begin{list}{}{\leftmargin=1.0em\itemsep=-0.1em}
              \item \textbf{a)} Activation of static/gimbaled thrusters with limited fuel
              \item \textbf{b)} Target linear \& angular acceleration
          \end{list}
    \item \textbf{Manipulator}
          \vspace{-0.25em}
          \begin{list}{}{\leftmargin=1.0em\itemsep=-0.1em}
              \item \textbf{a)} Operational Space Control~(OSC)
              \item \textbf{b)} Inverse Kinematics~(IK)
              \item \textbf{c)} Target joint positions
          \end{list}
    \item \textbf{End-Effector}
          \vspace{-0.25em}
          \begin{list}{}{\leftmargin=1.0em\itemsep=-0.1em}
              \item \textbf{a)} Target joint positions
              \item \textbf{b)} Target joint velocities
          \end{list}
\end{list}

\subsection{Support for Diverse Workflows}\label{subsec:diverse_workflows}

\begin{figure}[t]
    \centering
    \includegraphics[width=\linewidth]{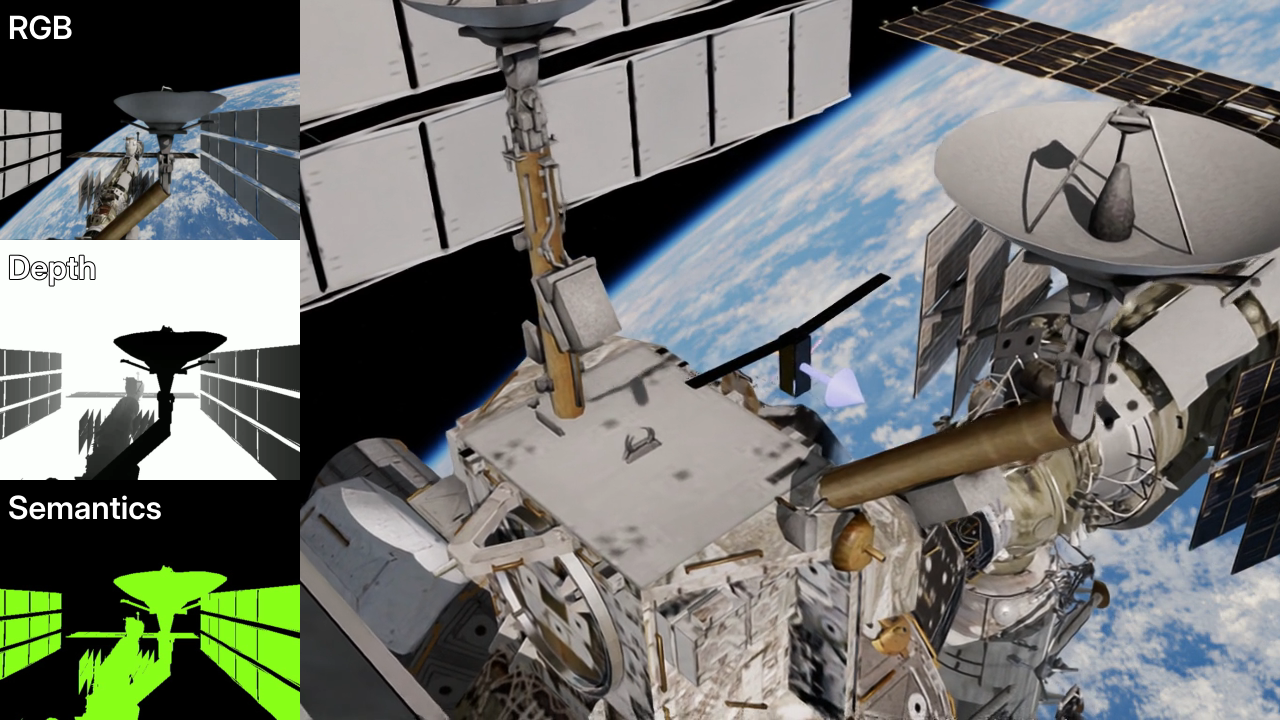}

    \vspace{0.25em}

    \includegraphics[width=\linewidth]{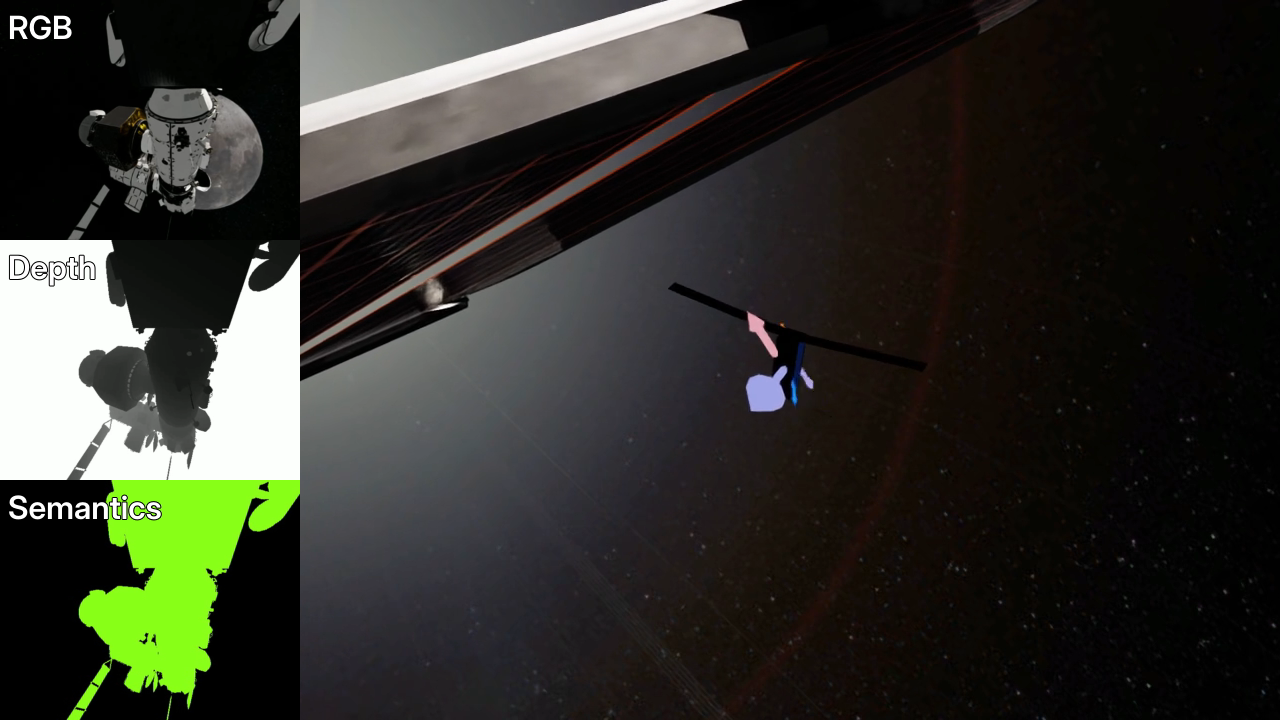}
    \caption{Orbital inspection scenario of SRB that provides synchronized RGB, depth, and segmentation streams from an onboard camera maneuvered around the ISS and Gateway.}
    \label{fig:orbital_inspection_visual}
    \vspace{-1.0em}
\end{figure}

While optimized for RL, the architecture of SRB is designed to support a broad spectrum of autonomy research and to bridge different development communities. A versatile teleoperation interface enables the collection of human demonstrations, providing the expert data essential for workflows in imitation learning and offline RL. For researchers focused on traditional robotics, a native ROS~2 interface~\cite{macenski2022ros2} serves as a critical bridge. It exposes all sensor data and control interfaces as standard topics and services, simplifying the integration of external software stacks. This allows for the validation of classical planning and control algorithms within diverse procedural environments. As demonstrated by the multi-modal sensor feeds for an orbital inspection scenario in Fig.~\ref{fig:orbital_inspection_visual}, the framework can also generate and stream rich visual data with ground truth annotations for perception-based systems. The core objective of each task is configurable via its reward function and termination conditions defined within a single function, which allows researchers to investigate fundamental questions in machine learning, such as the effects of reward shaping or curriculum learning strategies. This flexibility positions SRB not just as a benchmark, but as a versatile sandbox for comparative studies and multi-paradigm robotic development.

\subsection{Sim-to-Real Workflow}

A primary challenge in simulation-based robotics is the transfer of learned policies to physical hardware. SRB addresses this with a dedicated and streamlined workflow designed to minimize the engineering effort required for hardware deployment. This system provides a clear pathway from a trained policy checkpoint to execution on a physical robot.

The core of this workflow is an automated mechanism that, through runtime reflection, inspects a simulated Gymnasium environment and generates its lightweight, real-world counterpart. This hardware-specific environment does not depend on the simulation backend. Instead, it routes the actions and observations of an agent through a set of modular hardware interfaces. These interfaces are simple, self-contained components that abstract low-level communication protocols. For a typical ROS~2 platform, this involves linking agent actions to a ROS~2 publisher and observations to a subscriber. The generation process automatically handles critical details such as action space scaling and action rate to ensure consistency between the simulated and real domains.

Once this bridge is generated, the trained policy can be deployed for evaluation on the physical robot using the unified command-line interface. This process loads the policy checkpoint and the real-world environment, which in turn activates the specified hardware interfaces to control the robot and receive sensor data. The workflow also supports more advanced use cases such as fine-tuning a policy directly on the physical robot. Rewards and episode terminations can similarly be provided through the hardware interface system, for instance, via buttons or a communication middleware, to enable a complete on-robot training loop. While this capability is architecturally supported, an exploration of its effectiveness is beyond the scope of this work. The sim-to-real system is designed for extensibility, such that researchers can add support for new sensors or actuators by implementing a new hardware interface that is automatically registered with SRB. This modular design ensures that the workflow can be adapted to a wide range of custom robotic platforms, providing a versatile and robust solution for bridging the gap between simulation and reality.

\section{Benchmark Tasks}
\label{sec:tasks}

Building upon the SRB framework, we introduce a suite of benchmark tasks designed to evaluate robotic capabilities across a wide range of mission-relevant challenges in space. As summarized in Table~\ref{tab:srb_tasks}, these tasks are organized into three primary categories: \textbf{mobile robotics}, \textbf{manipulation}, and \textbf{mobile manipulation}. Each task is formulated to test specific aspects of autonomous operation, from fundamental motor control and reactive planning to long-horizon contact-rich interaction.

\begin{table*}[t]
    \centering
    \caption{\textsc{Overview of the Standard SRB Benchmark Tasks.}}
    \label{tab:srb_tasks}
    \begin{tabular}{@{}llc@{}}
        \toprule
        \textbf{Task ID}                                   & \textbf{Objective}                                                                          & \textbf{Robot Morphology}    \\
        \midrule
        \multicolumn{3}{@{}l@{}}{\textit{\textbf{Mobile Robotics}}}                                                                                                                     \\
        \quad\texttt{landing}                              & Descend and safely land with limited fuel $\vert$ focus: terminal control                   & Spacecraft                   \\
        \quad\texttt{rendezvous}                           & Approach and match the state of a tumbling object $\vert$ focus: precision control          & Spacecraft                   \\
        \quad\texttt{orbital\_evasion}                     & Maneuver to avoid dynamic obstacles $\vert$ focus: reactive planning                        & Spacecraft                   \\
        \quad\texttt{velocity\_tracking}                   & Follow dynamic velocity commands $\vert$ focus: low-level motor control                     & Wheeled                      \\
        \quad\quad\texttt{$\hookrightarrow$ locomotion\_*} & $\hookrightarrow$ Variant for legged systems $\vert$ extra focus: whole-body coordination   & Legged/Humanoid              \\
        \quad\texttt{waypoint\_navigation}                 & Track a dynamic waypoint $\vert$ focus: long-term trajectory tracking                       & Wheeled                      \\
        \quad\quad\texttt{$\hookrightarrow$ locomotion\_*} & $\hookrightarrow$ Variant for legged systems $\vert$ extra focus: high-level planning       & Legged/Humanoid              \\
        \quad\quad\texttt{$\hookrightarrow$ orbital\_*}    & $\hookrightarrow$ Variant for spacecraft $\vert$ extra focus: underactuated 3D dynamics     & Spacecraft                   \\
        \midrule
        \multicolumn{3}{@{}l@{}}{\textit{\textbf{Manipulation (Fixed Base)}}}                                                                                                           \\
        \quad\texttt{debris\_capture}                      & Capture and stabilize a tumbling debris $\vert$ focus: dynamic object grasping              & Manipulator                  \\
        \quad\texttt{sample\_collection}                   & Grasp a domain-specific sample $\vert$ focus: robust grasping of novel geometries           & Manipulator                  \\
        \quad\quad\texttt{$\hookrightarrow$ multi\_*}      & $\hookrightarrow$ Variant with multiple samples $\vert$ extra focus: generalization         & Manipulator                  \\
        \quad\texttt{excavation}                           & Excavate and lift granular media $\vert$ focus: interaction with deformable materials       & Manipulator                  \\
        \quad\texttt{peg\_in\_hole}                        & Pick up and precisely insert a peg into its hole $\vert$ focus: contact-rich interaction    & Manipulator                  \\
        \quad\quad\texttt{$\hookrightarrow$ multi\_*}      & $\hookrightarrow$ Variant with multiple assemblies $\vert$ extra focus: high-level planning & Manipulator                  \\
        \quad\texttt{screwdriving}                         & Fasten a bolt into a threaded hole $\vert$ focus: tool use and force modulation             & Manipulator                  \\
        \quad\texttt{solar\_panel\_assembly}               & Assemble a structure via a sequence of insertions $\vert$ focus: long-horizon planning      & Manipulator                  \\
        \midrule
        \multicolumn{3}{@{}l@{}}{\textit{\textbf{Mobile Manipulation}}}                                                                                                                 \\
        \quad\texttt{mobile\_debris\_capture}              & Approach and capture a tumbling debris $\vert$ focus: whole-body coordination               & Spacecraft + Manipulator     \\
        \quad\texttt{mobile\_excavation}                   & Excavate and store granular media $\vert$ focus: long-horizon planning                      & Wheeled/Legged + Manipulator \\
        \bottomrule
    \end{tabular}
    \vspace{-1.0em}
\end{table*}

\subsection{Design Philosophy}

The design of our benchmark tasks prioritizes the evaluation of complete, mission-relevant skills over many narrow, isolated sub-problems. For example, rather than providing separate tasks for standing, walking forward, and turning, the \texttt{locomotion\_velocity\_tracking} task requires an agent to learn a single versatile policy that can respond to continuous high-level velocity commands. Similarly, manipulation tasks focus on the entire objective. The \texttt{peg\_in\_hole} assembly task requires the agent to perform the full sequence of picking up a peg and precisely inserting it, rather than focusing solely on the final insertion or a simplified reaching sub-task. This holistic approach produces more practical and generalizable skills that can be composed into complex, long-horizon behaviors.

To facilitate broad applicability, the core task logic can be shared and adapted across different domains and robot morphologies. For instance, the fundamental challenge of navigating to a dynamic waypoint is presented in several distinct variants. The \texttt{waypoint\_navigation} task is designed for wheeled rovers on planetary surfaces, while the \texttt{locomotion\_waypoint\_navigation} variant extends this challenge for the unique dynamics of legged systems. Similarly, the \texttt{orbital\_waypoint\_navigation} variant reformulates the objective for spacecraft in a microgravity environment, requiring full 6DOF navigation with underactuated dynamics. This modular design enables direct studies into how different robot morphologies and physical domains affect the learning of a common skill.

Furthermore, research into generalization is enabled by designing SRB tasks to be as configurable as possible beyond the standard benchmark setup. The \texttt{sample\_collection} task, for instance, can switch between using a single object from a static dataset or a set of multiple diverse objects sampled from a procedural distribution. This feature provides researchers with a direct mechanism to control environmental diversity and systematically measure its impact on policy robustness.

\subsection{Observations}

To support a wide range of learning paradigms, from traditional state-based RL to end-to-end visuomotor control, every task is formulated as a partially observable Markov decision process~(POMDP) with a multi-modal observation space. Observations are structured into a dictionary of tensors, allowing an agent to flexibly use different information sources. This structured approach clearly separates privileged simulation information from achievable sensor data, enabling fair and realistic evaluation of different policies. Furthermore, it directly contributes to the sim-to-real mechanism by isolating the subset of observations that would be available on a physical robot. The observation modalities are grouped into four primary categories:
\begin{list}{}{\leftmargin=0em\itemsep=0em}
    \item \textbf{State:} Privileged simulation information (transformations\footnote{We encode all agent-facing rotations using the 6D representation~\cite{zhou2019continuity}}, velocities, accelerations, forces, contacts, task-specific data)
    \item \textbf{Proprioception:} Internal measurements (kinematic state, IMU readings, remaining fuel)
    \item \textbf{Visual:} Rendered images (RGB images, depth maps, normals, segmentation masks)
    \item \textbf{Commands:} High-level control signals (target velocity, relative waypoint pose)
\end{list}
To clearly distinguish between learning paradigms and optimize performance, tasks that utilize visual sensors are purposefully registered as separate Gymnasium environments with a \texttt{\_visual} suffix. This design choice makes the intended input modality explicit, and it also ensures that non-visual tasks can run more efficiently without the computationally expensive rendering pipelines and sensor processing plugins.

These categories are further divided based on whether their dimensionality is fixed or varies with the robot morphology. This distinction is a deliberate design choice to facilitate research at scale. On a practical level, it simplifies data collection and batching across experiments with different robots, as fixed-size tensors can be handled uniformly. More importantly, this design enables research into cross-morphology policy learning. By isolating morphology-invariant observations, the framework provides a clear pathway for training generalist agents that can transfer across different robotic platforms. For example, IMU data and end-effector pose are provided as fixed-size vectors, while the full kinematic state is variable.

\subsection{Rewards}

To guide the learning process on these challenges, each task implements a composite reward function. These functions are structured as a weighted sum of multiple terms that encourage incremental progress towards the primary objective while penalizing undesirable behaviors. This design provides a continuous and informative learning signal, which is crucial for sample efficiency when dealing with complex high-dimensional problems. The provided reward functions are intended to produce competent baseline behaviors that can serve as a strong starting point for further research.

The reward components are shaped to be mission-relevant, often combining terms that guide an agent through a logical task progression with penalties that enforce safety and efficiency. Positive terms frequently reward sequential sub-goals, such as approaching a target, correctly orienting an end-effector, successfully interacting with an object, and achieving a stable final state. These are balanced by penalties for actions that could compromise the mission or damage the hardware. Common penalties include terms for high-frequency actions, excessive joint-space jerk, high contact forces, energy or propellant consumption, and creating environmental hazards. For example, the \texttt{landing} task reward prioritizes minimizing velocity just before touchdown to ensure a soft landing, while the \texttt{excavation} task balances maximizing the scooped volume of regolith with minimizing dust generation. Some tasks, like \texttt{screwdriving}, also incorporate sparse success bonuses for completing discrete sub-objectives to enable research into both dense and sparse reward settings.

However, we acknowledge that reward engineering is a nuanced and often subjective aspect of robot learning. The provided handcrafted functions, while effective for establishing baselines, may unintentionally bias an agent towards preconceived solutions and potentially limit the discovery of more novel or efficient strategies. As mentioned in Section~\ref{subsec:diverse_workflows}, SRB therefore formulates the entire reward function for each task inside a single function that can be modified or replaced. This design choice allows researchers to systematically study the effects of reward shaping or even explore alternative learning paradigms that reduce the reliance on manual reward design, such as learning from human demonstrations, preferences, or intrinsic motivation.

\section{Baselines and Performance}
\label{sec:baselines}

To establish foundational metrics for SRB, this section presents baseline results focusing on two key aspects: the computational efficiency of the simulation framework and the performance of standard RL algorithms across the suite of benchmark tasks. These results quantify the capabilities and provide a clear reference point for future research.

\subsection{Simulation Throughput}

High-throughput parallel simulation is a fundamental requirement for modern robot learning research, as it enables the efficient collection of the vast and diverse data required by high-capacity models. To quantify the capabilities of the framework in this regard, we evaluate the simulation throughput of all SRB tasks. Performance is measured as the aggregate number of simulation steps processed per second across all parallel environments. The evaluation was conducted over a runtime of five minutes on a workstation with an AMD Ryzen~9~7950X CPU and an NVIDIA~RTX~4090 GPU, using the default task configurations with randomly sampled actions.

\begin{figure}[t]
    \centering
    \includegraphics[width=\linewidth]{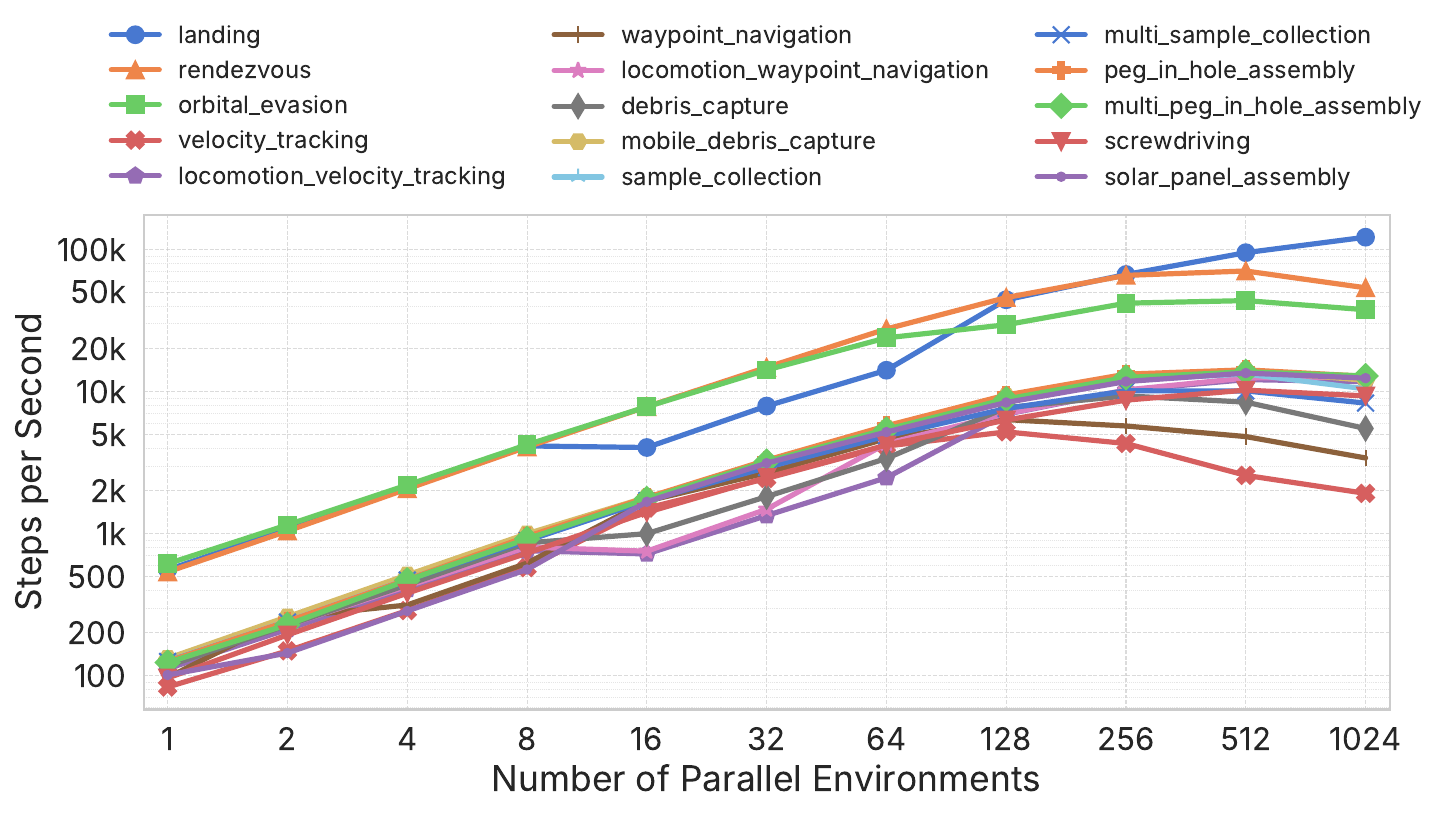}
    \caption{The aggregate throughput of SRB tasks with respect to the number of parallel environment instances.}
    \label{fig:simulation_performance}
    \vspace{-1.0em}
\end{figure}

The results, presented in Fig.~\ref{fig:simulation_performance}, demonstrate the high-throughput capabilities of SRB. Throughput varies across task categories, primarily influenced by the complexity of the physics simulation. Computationally lighter tasks like \texttt{landing} achieve over 100k steps per second with 1024 parallel environments. Most locomotion and manipulation tasks, which involve more complex contact dynamics, plateau at approximately 15k steps per second with 512 parallel environments. In contrast, the \texttt{excavation} task, which uses a high-fidelity particle simulation for granular media, is significantly more computationally intensive and maintains a constant throughput of around 40 steps per second regardless of the environment count. This high performance across a wide range of tasks dramatically accelerates the research lifecycle, from hyperparameter tuning to large-scale policy training.

\subsection{Reinforcement Learning Baselines}

To provide a reference for future work and to analyze the difficulty of the benchmark tasks, we evaluate three widely adopted RL algorithms that represent distinct learning paradigms: Proximal Policy Optimization~(PPO)~\cite{schulman2017proximal}, a stochastic on-policy algorithm; Twin Delayed Deep Deterministic Policy Gradient~(TD3)~\cite{fujimoto2018addressing}, a deterministic off-policy algorithm; and DreamerV3~\cite{hafner2025mastering}, a model-based agent trained via latent imagination. For PPO and TD3, we use the implementations of Stable-Baselines3~\cite{raffin2021stable}.

\begin{figure*}[t]
    \centering
    \captionsetup[subfigure]{aboveskip=-1.0pt,belowskip=0.0pt}
    \subcaptionbox*{\scriptsize\texttt{\textbf{landing}}}{\includegraphics[width=0.24\linewidth]{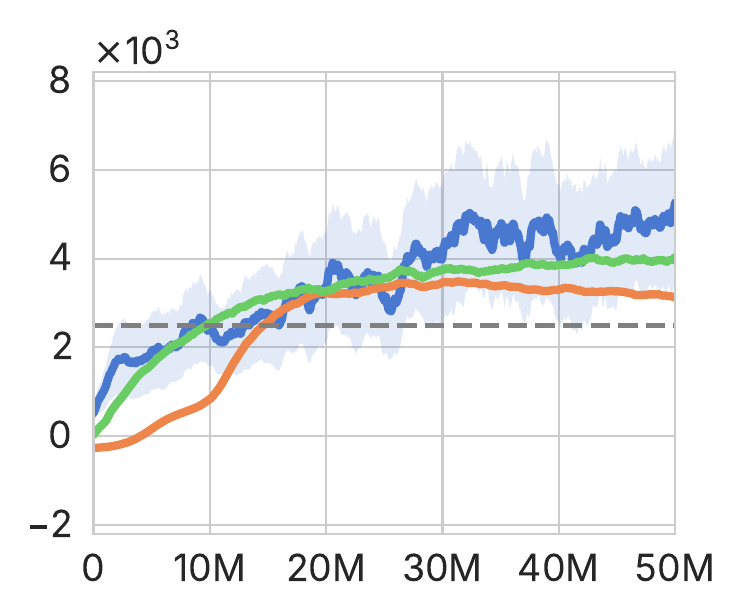}}
    \subcaptionbox*{\scriptsize\texttt{\textbf{rendezvous}}}{\includegraphics[width=0.24\linewidth]{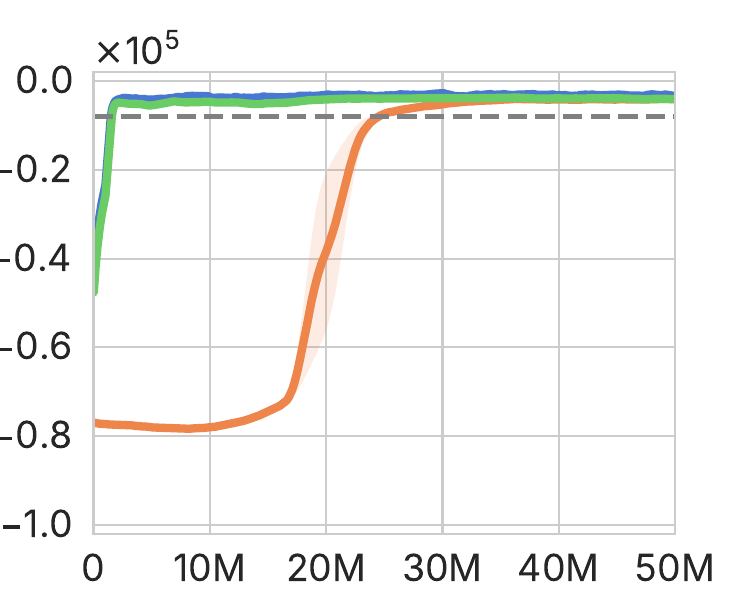}}
    \subcaptionbox*{\scriptsize\texttt{\textbf{orbital\_evasion}}}{\includegraphics[width=0.24\linewidth]{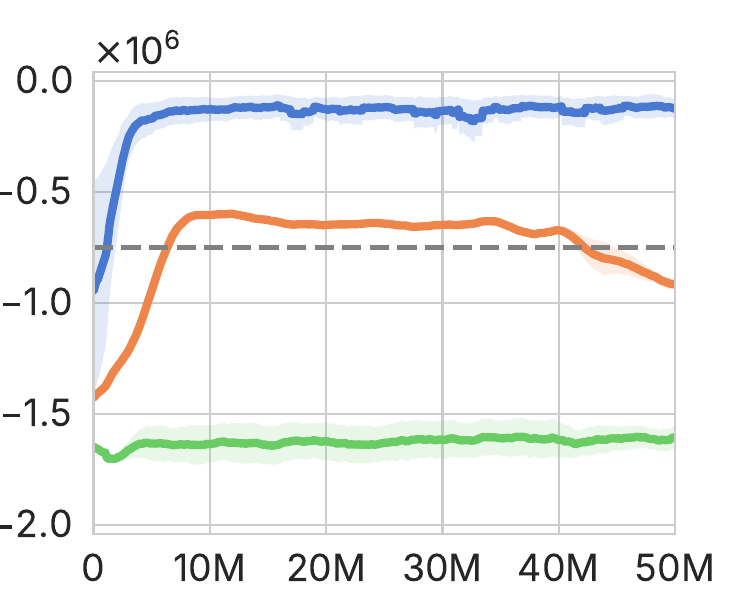}}
    \subcaptionbox*{\scriptsize\texttt{\textbf{debris\_capture}}}{\includegraphics[width=0.24\linewidth]{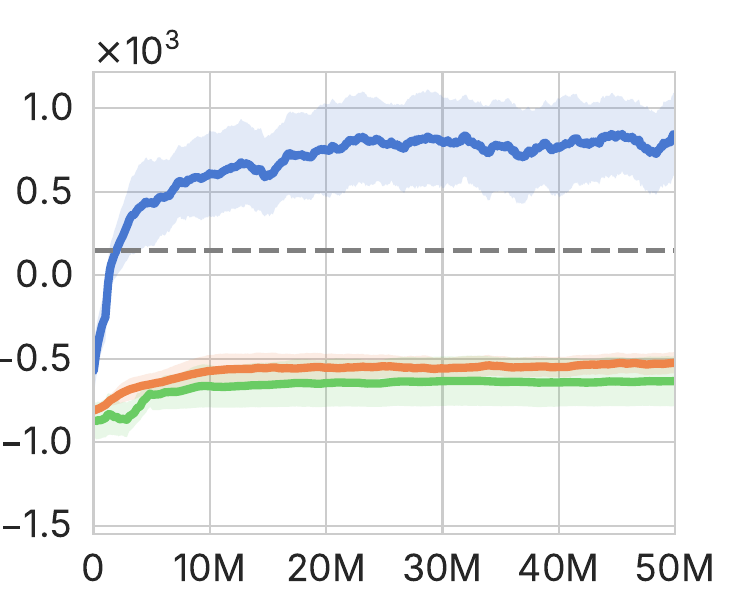}}
    \subcaptionbox*{\scriptsize\texttt{\textbf{velocity\_tracking}}}{\includegraphics[width=0.24\linewidth]{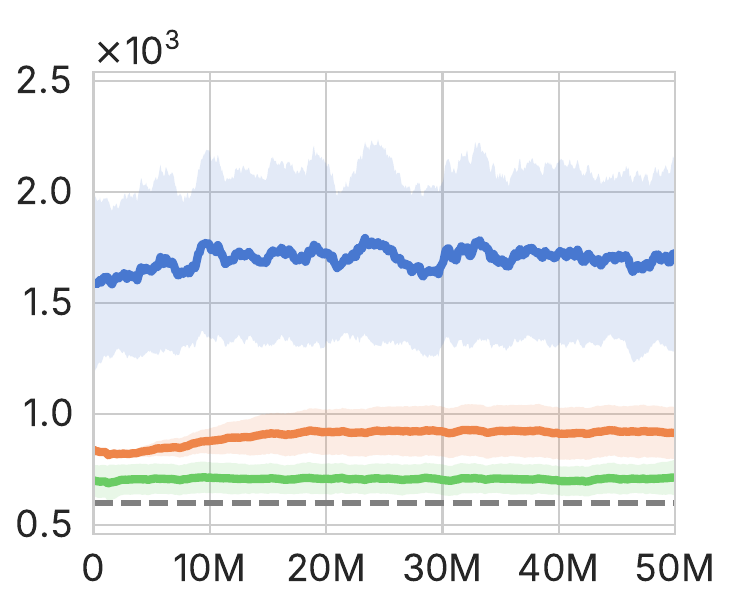}}
    \subcaptionbox*{\fontsize{6.9}{7.4175}\texttt{\textbf{locomotion\_velocity\_tracking}}}{\includegraphics[width=0.24\linewidth]{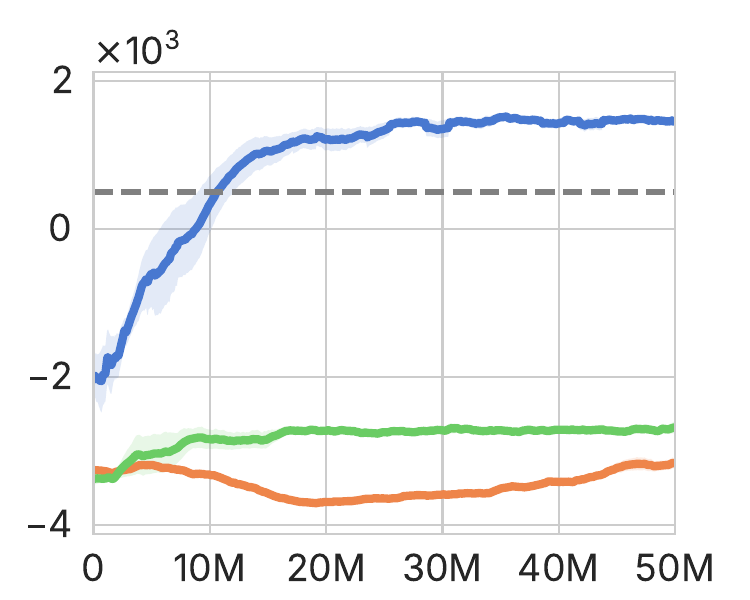}}
    \subcaptionbox*{\scriptsize\texttt{\textbf{waypoint\_navigation}}}{\includegraphics[width=0.24\linewidth]{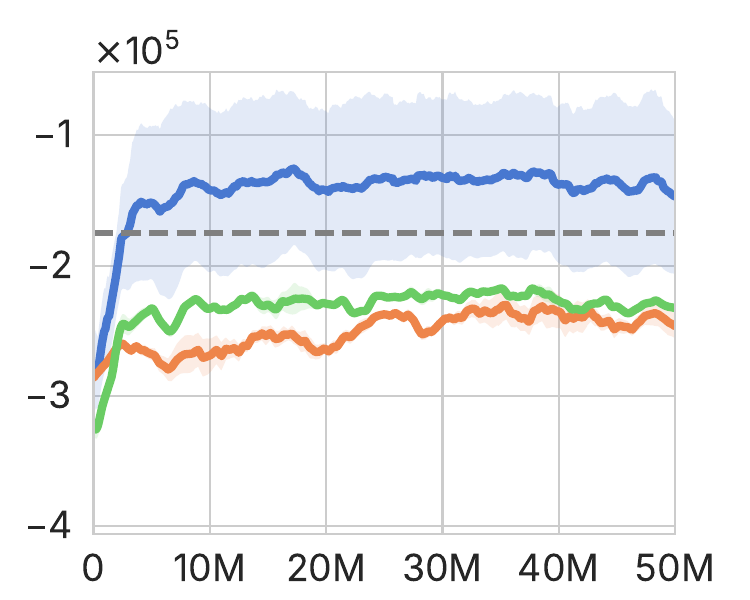}}
    \subcaptionbox*{\fontsize{6.9}{7.4175}\selectfont\texttt{\textbf{locomotion\_waypoint\_navigation}}}{\includegraphics[width=0.24\linewidth]{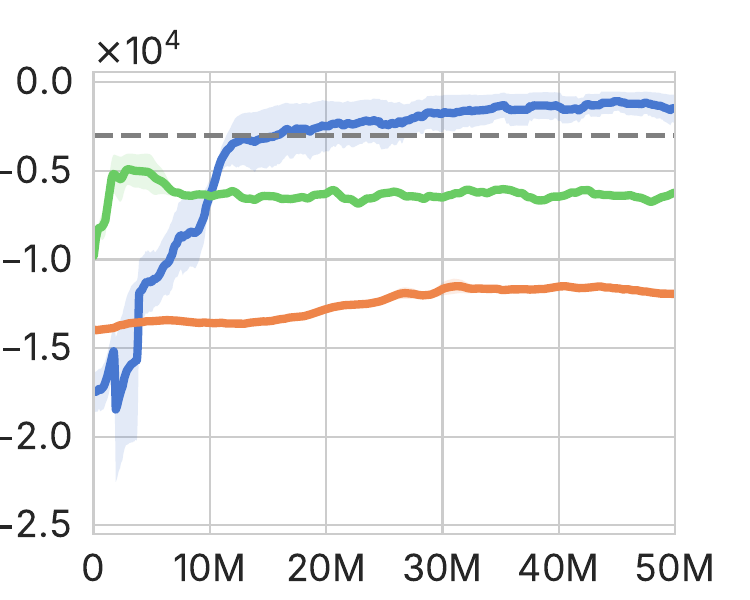}}
    \subcaptionbox*{\scriptsize\texttt{\textbf{sample\_collection}}}{\includegraphics[width=0.24\linewidth]{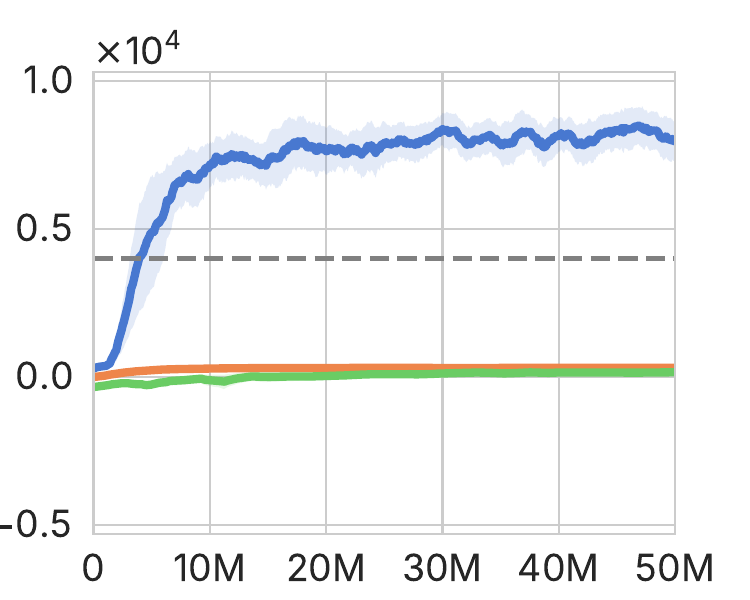}}
    \subcaptionbox*{\scriptsize\texttt{\textbf{excavation}}}{\includegraphics[width=0.24\linewidth]{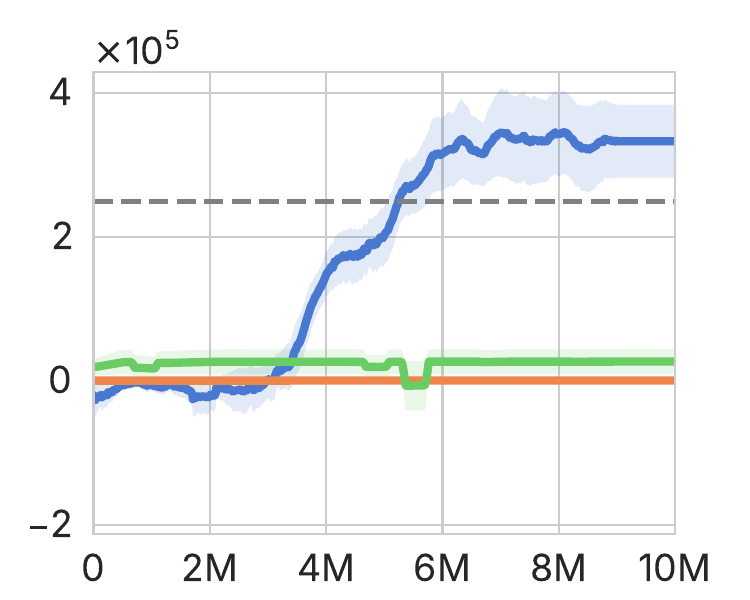}}
    \subcaptionbox*{\scriptsize\texttt{\textbf{peg\_in\_hole}}}{\includegraphics[width=0.24\linewidth]{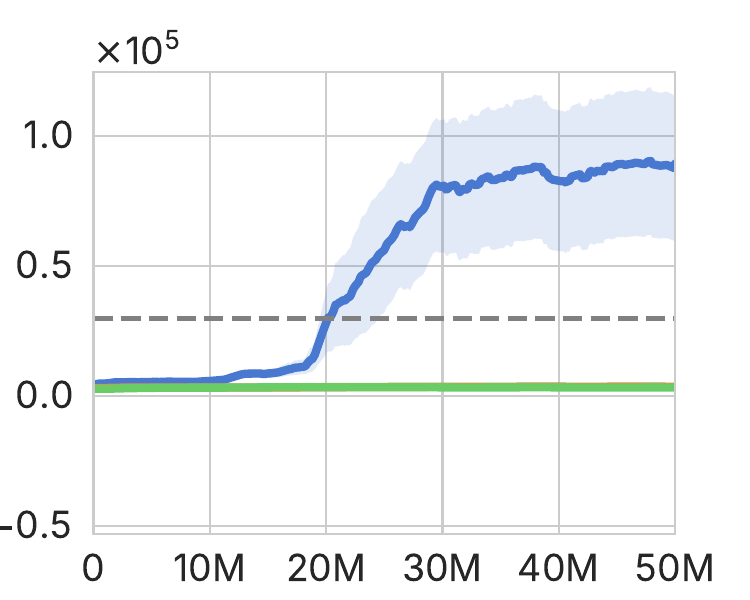}}
    \subcaptionbox*{\scriptsize\texttt{\textbf{screwdriving}}}{\includegraphics[width=0.24\linewidth]{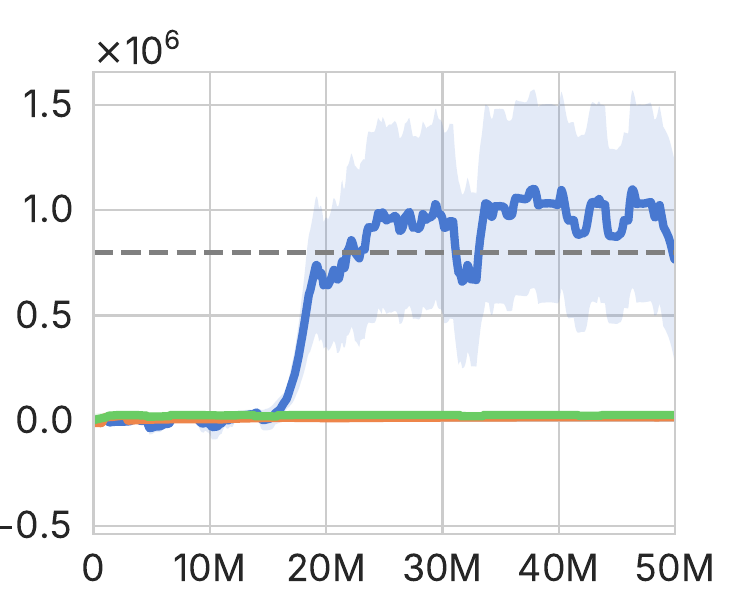}}
    \includegraphics[width=0.24\linewidth]{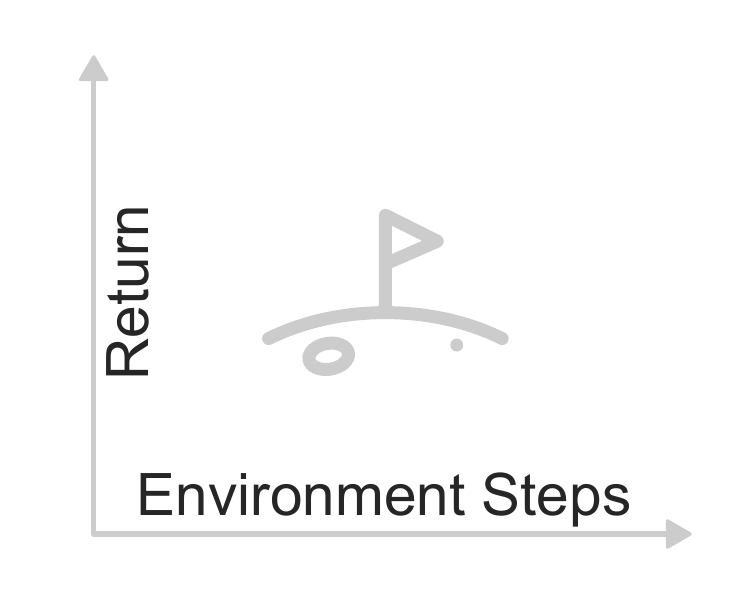}
    \subcaptionbox*{\scriptsize\texttt{\textbf
            {mobile\_debris\_capture}}}{\includegraphics[width=0.24\linewidth]{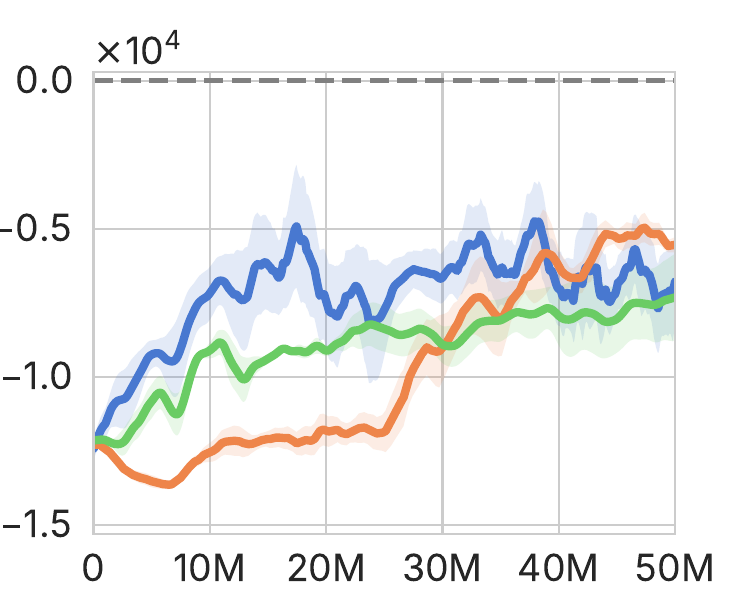}}
    \subcaptionbox*{\scriptsize\texttt{\textbf{solar\_panel\_assembly}}}{\includegraphics[width=0.24\linewidth]{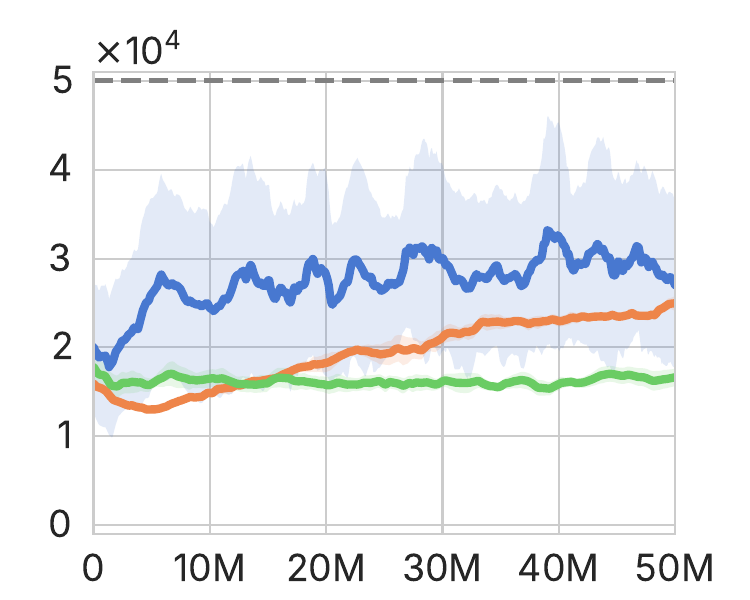}}
    \includegraphics[width=0.24\linewidth]{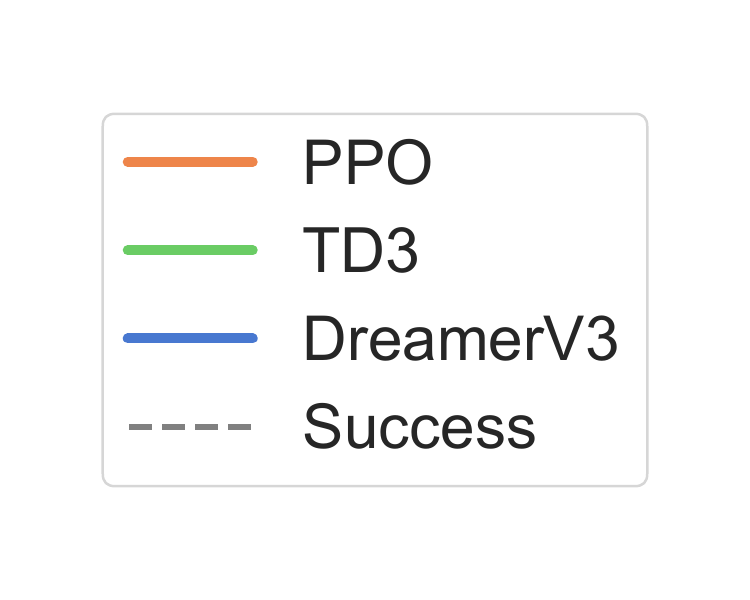}
    \caption{Learning curves of RL baselines, averaged over three random seeds. Shaded regions represent the standard deviation. The dashed lines in plots with a defined success condition qualitatively indicate the threshold for consistent task success.}
    \label{fig:baseline_learning_curves}
    \vspace{-1.0em}
\end{figure*}

All agents were trained using the default task configurations with state-based observations on a single NVIDIA~RTX~4090 GPU. We used 512 parallel environments for most tasks, and 16 for the computationally demanding \texttt{excavation} task. The primary hyperparameters, kept consistent across all tasks for a fair comparison, are detailed in Appendix~\ref{app:hyperparameters}. The learning curves, averaged over three random seeds, are presented in Fig.~\ref{fig:baseline_learning_curves}. The results show that DreamerV3 consistently demonstrates superior sample efficiency, converging to higher episodic returns and success rates across nearly all tasks. However, this performance comes at a greater computational cost in terms of wall-clock training time. With our setup and 32 updates per environment step, DreamerV3 required on average 5.3\texttimes\ longer to train than PPO for the same number of environment steps. Notably, tasks requiring long-horizon planning and high-dimensional coordination, such as the \texttt{solar\_panel\_assembly} and \texttt{mobile\_debris\_capture}, remain unsolved by all three algorithms. This highlights these tasks as significant open challenges for the research community, providing a valuable testbed for the development of new algorithms in hierarchical learning and complex motor control.

\section{Experimental Case Studies}
\label{sec:experiments}

To validate our proposed procedural paradigm and demonstrate the utility of SRB for advancing space robotics research, we present a series of experimental case studies. These studies go beyond baseline performance to showcase how unique features of SRB can be leveraged to tackle key challenges. We begin with a validation of the entire workflow through successful zero-shot sim-to-real transfer. We then present supporting case studies that highlight the capabilities of SRB for investigating the principles of generalization and developing adaptive control strategies.

\subsection{Zero-Shot Sim-to-Real Transfer}

The primary validation of our paradigm is a successful zero-shot transfer for the mission-relevant challenge of dynamic waypoint tracking with a rover on deformable granular media. This experiment directly addresses the critical questions of hardware validation and the sim-to-real gap.

\paragraph{Experimental Setup}
All real-world validation was performed inside a lunar-analogue testbed with basalt gravel using Leo Rover, a four-wheeled skid-steer platform. Ground-truth localization was provided by an external OptiTrack motion capture system. We trained PPO, PPO with LSTM, TD3, and DreamerV3 agents solely inside SRB using 512 parallel environment instances, which were then deployed zero-shot to the physical rover.

\begin{figure}[t]
    \centering
    \includegraphics[width=\linewidth]{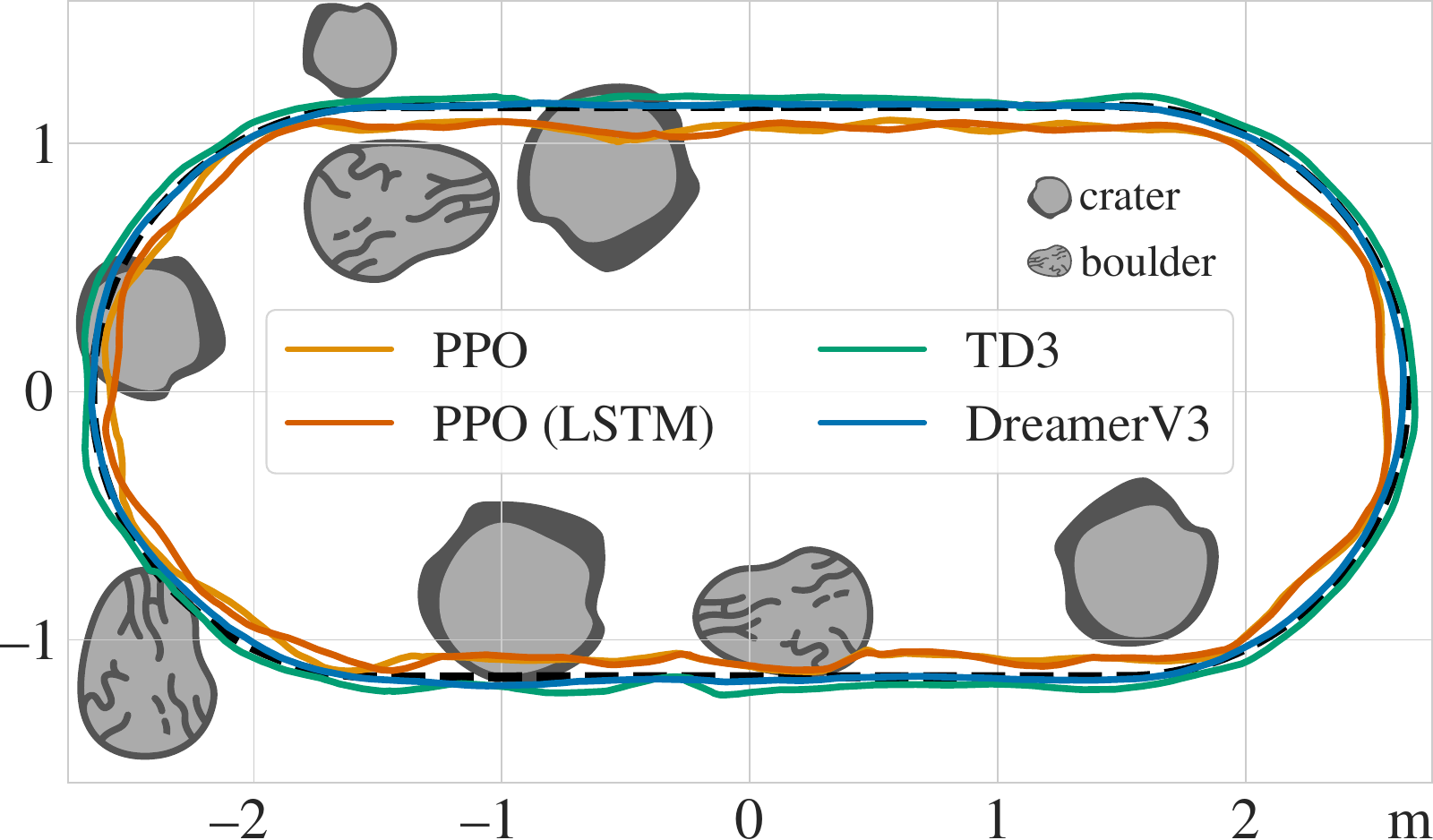}

    \vspace{0.25em}

    \includegraphics[width=\linewidth]{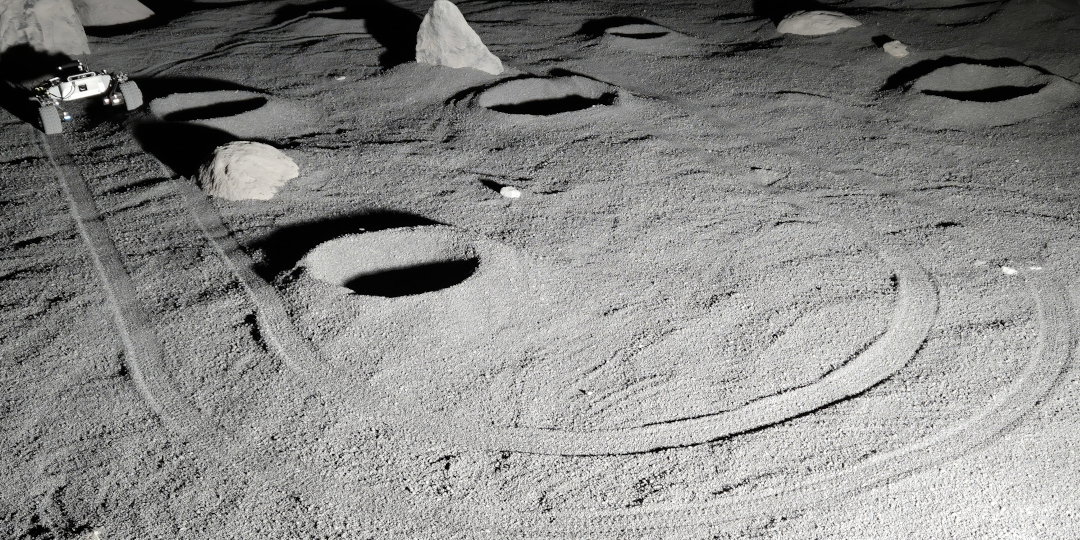}
    \caption{Sim-to-real trajectory tracking of different RL agents.}
    \label{fig:sim2real_trajectories}
    \vspace{-1em}
\end{figure}

\paragraph{Algorithmic comparison}
As shown in Table~\ref{tab:sim2real_algos} and Fig.~\ref{fig:sim2real_trajectories}, the model-based DreamerV3 agent demonstrated vastly superior real-world performance, achieving a substantially lower Average Tracking Error~(ATE) across all velocity profiles. Based on this result, DreamerV3 was selected for all subsequent experiments.

\begin{table}[ht]
    \centering
    \caption{\textsc{Zero-Shot Sim-to-Real ATE Performance of Different RL Algorithms on the Dynamic Waypoint Tracking Task.}}
    \label{tab:sim2real_algos}
    \addtolength{\tabcolsep}{-0.175em}
    \begin{tabular}{@{}rcccc@{}}
        \toprule
        \textbf{Velocity} & \textbf{PPO}             & \textbf{PPO (LSTM)}      & \textbf{TD3}             & \textbf{DreamerV3}               \\
        \midrule
        5  cm/s           & 13.2 cm / 7.8\textdegree & 11.4 cm / 4.8\textdegree & 12.6 cm / 8.5\textdegree & \textbf{2.3 cm / 1.7\textdegree} \\
        15 cm/s           & 13.7 cm / 8.6\textdegree & 11.2 cm / 8.1\textdegree & 11.6 cm / 6.2\textdegree & \textbf{3.3 cm / 1.9\textdegree} \\
        25 cm/s           & 14.8 cm / 8.7\textdegree & 12.9 cm / 9.9\textdegree & 13.1 cm / 9.1\textdegree & \textbf{3.6 cm / 2.3\textdegree} \\
        \bottomrule
    \end{tabular}
\end{table}

\paragraph{Role of Procedural Diversity} To isolate the effect of environmental diversity on sim-to-real transfer, we compared DreamerV3 agents trained under two regimes: \textit{Static}, where all parallel instances shared a single procedurally generated terrain; and \textit{Procedural}, where each of the 512 instances used a unique terrain. While both agents performed well in simulation, their real-world capabilities differed dramatically. The policy trained with diversity proved far superior. At a speed of 15~cm/s, the procedural agent maintained a low ATE of 3.3~cm and 1.9\textdegree, whereas the error of the statically trained agent was significantly higher at 4.2~cm and 6.8\textdegree. This pattern of improved precision and stability was consistent across all tested velocities. Overall, the procedural agent achieved a 21\% lower location ATE and a 72\% reduction in orientation ATE compared to its statically trained counterpart. This result provides strong evidence that the procedural paradigm is a practical and effective workflow for creating policies that are robust to the unknown variations of the physical world.

\subsection{General Principle of Diversity}

Having established the critical role of procedural diversity for successful sim-to-real transfer, we conducted further simulation-based experiments to analyze its universal impact across three distinct axes of diversity: the environment, interactive objects, and the robot morphology. Agents were again trained in either \textit{Static} or \textit{Procedural} conditions and then evaluated on their zero-shot transfer performance to out-of-distribution scenarios.

As summarized in Table~\ref{tab:generalization_results}, the procedural paradigm consistently yields more generalizable policies, regardless of the source of diversity. For environmental diversity in tasks like locomotion over varied terrains, procedural training resulted in a slight advantage. The benefit was even more pronounced for object interaction. In the \texttt{sample\_collection} task, the performance of the statically trained agent collapsed by nearly 50\% when faced with unseen procedural rocks. In contrast, the procedural agent maintained its high performance, demonstrating a robust ability to generalize to novel object geometries.

Finally, we tested the paradigm on the most challenging axis via the morphology of the robot that the agent controls. In the \texttt{excavation} task, we compared an agent trained on a single tool geometry against a more general agent trained on a procedural distribution of 16 unique tools. The specialist, while proficient with its one tool, failed completely to adapt to novel, held-out tool geometries. At the same time, the generalist agent successfully generalized its excavation skill, achieving a nearly tenfold increase in performance. This cohesive result across all three axes of diversity confirms that systematically exposing an agent to procedural variation, whether in the world or its own embodiment, is a powerful and universally applicable method for forging robust policies.

\begin{table}[ht]
    \centering
    \caption{\textsc{Normalized Episodic Return of Zero-Shot Transfer to Out-of-Distribution Scenarios.}}
    \label{tab:generalization_results}
    \addtolength{\tabcolsep}{-0.175em}
    \begin{tabular}{@{}rlcc@{}}
        \toprule
        \textbf{Task}                        & \textbf{Diversity} & \textbf{Static} & \textbf{Procedural} \\
        \midrule
        \texttt{loco*\_waypoint\_navigation} & Environment        & 1.00            & \textbf{1.03}       \\
        \texttt{sample\_collection}          & Object             & 1.00            & \textbf{1.91}       \\
        \texttt{excavation}                  & Morphology         & 1.00            & \textbf{9.06}       \\
        \bottomrule
    \end{tabular}
    \vspace{-1.0em}
\end{table}

\subsection{Learning Compliant Manipulation}

\begin{figure}[ht]
    \vspace{-0.5em}
    \centering
    \includegraphics[width=\linewidth]{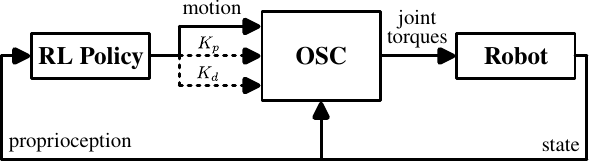}
    \caption{The action space of the OSC agent combines motion commands with stiffness and damping gains to achieve learned adaptive compliance.}
    \label{fig:osc}
\end{figure}

Rigid control strategies are often brittle and unsafe for the unpredictable, contact-rich scenarios common in space. To address this, the modular actuation models of SRB provide a testbed for investigating advanced alternatives like OSC~\cite{khatib1987osc}. We conducted a systematic comparison of a conventional rigid IK controller against an adaptive OSC policy that learned to dynamically modulate its own stiffness and damping gains in addition to providing motion commands, as illustrated in Fig.~\ref{fig:osc}. As demonstrated across a spectrum of manipulation tasks, learning to control compliance is a critical capability for achieving robust, safe, and effective physical interaction.

The results, summarized in Table~\ref{tab:osc_results}, reveal a pattern of improvement across different types of interaction challenges. For some delicate orbital operations like \texttt{debris\_capture}, where minimizing vibrations is paramount, the adaptive agent achieved a 58\% reduction in motion jerkiness. For high-precision assembly in the \texttt{peg\_in\_hole} task, the ability of the agent to compliantly negotiate contact states resulted in a higher final success rate. Finally, for high-force interaction with deformable media in the \texttt{excavation} task, learned compliance was essential for managing unpredictable forces, allowing the agent to excavate more material with 56\% less jerk. This cohesive result demonstrates that SRB serves as an effective platform for developing and validating advanced control methods that are fundamentally safer and more effective for complex physical interaction in space.

\begin{table}[ht]
    \centering
    \caption{\textsc{Performance of DreamerV3 using IK and OSC controllers across SRB Manipulation Tasks.}}
    \label{tab:osc_results}
    \begin{tabular}{@{}l|cc|cc@{}}
        \toprule
        \multirow{2}{*}{\textbf{Task}} & \multicolumn{2}{c|}{\textbf{Rigid IK}} & \multicolumn{2}{c}{\textbf{Adaptive OSC}}                                   \\
                                       & \textit{Return}                        & \textit{Jerk}                             & \textit{Return} & \textit{Jerk} \\
        \midrule
        \texttt{debris\_capture}       & \textbf{1.00}                          & 1.00                                      & 0.97            & \textbf{0.42} \\
        \texttt{peg\_in\_hole}         & 1.00                                   & 1.00                                      & \textbf{1.10}   & \textbf{0.50} \\
        \texttt{excavation}            & 1.00                                   & 1.00                                      & \textbf{1.08}   & \textbf{0.44} \\
        \bottomrule
    \end{tabular}
    \vspace{-1.0em}
\end{table}

\subsection{End-to-End Visuomotor Control}

\begin{figure}[ht]
    \centering
    \subcaptionbox{\texttt{landing}}{%
        \includegraphics[width=0.48\linewidth]{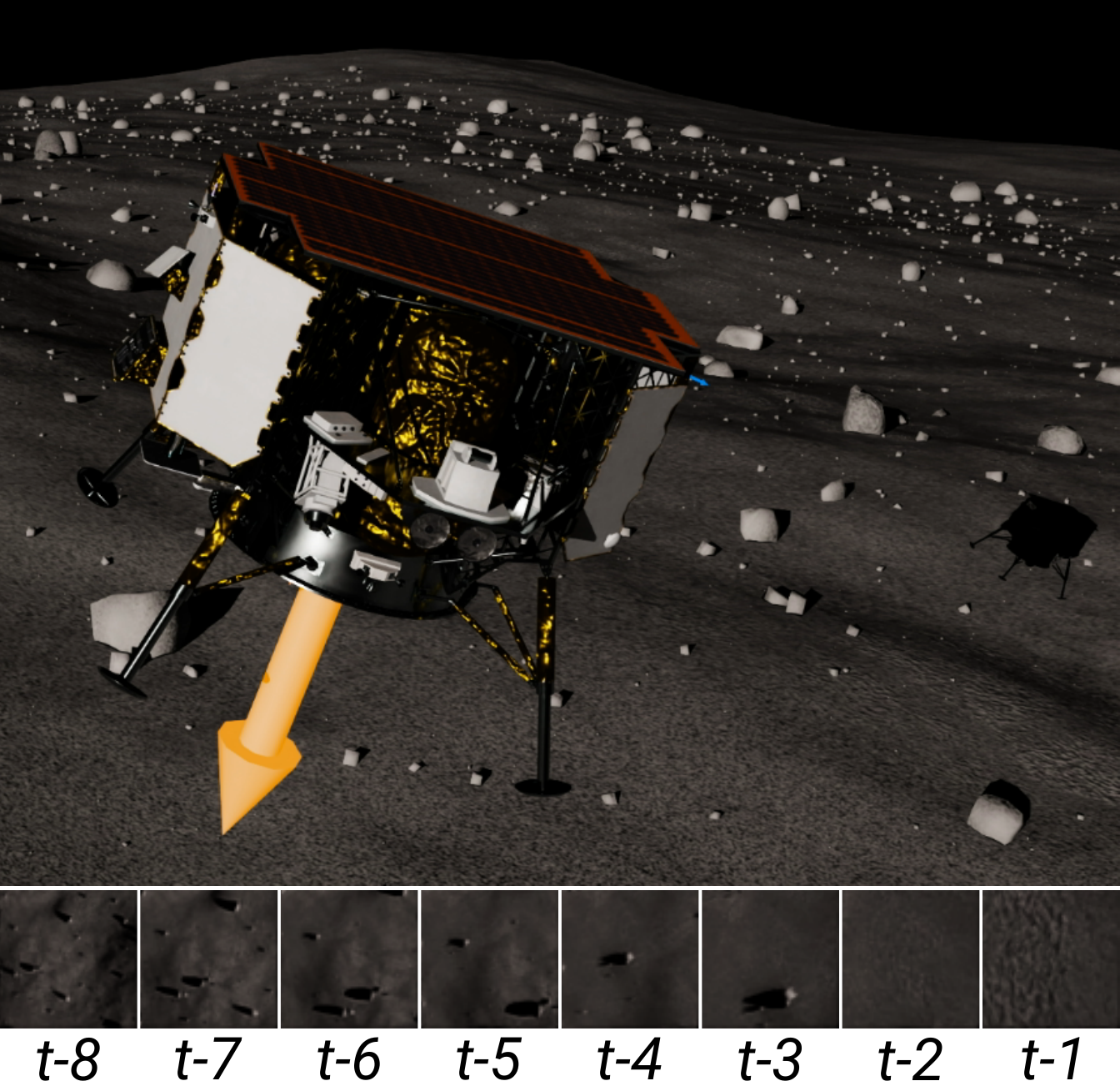}
    }
    \hfill
    \subcaptionbox{\texttt{peg\_in\_hole}}{%
        \includegraphics[width=0.48\linewidth]{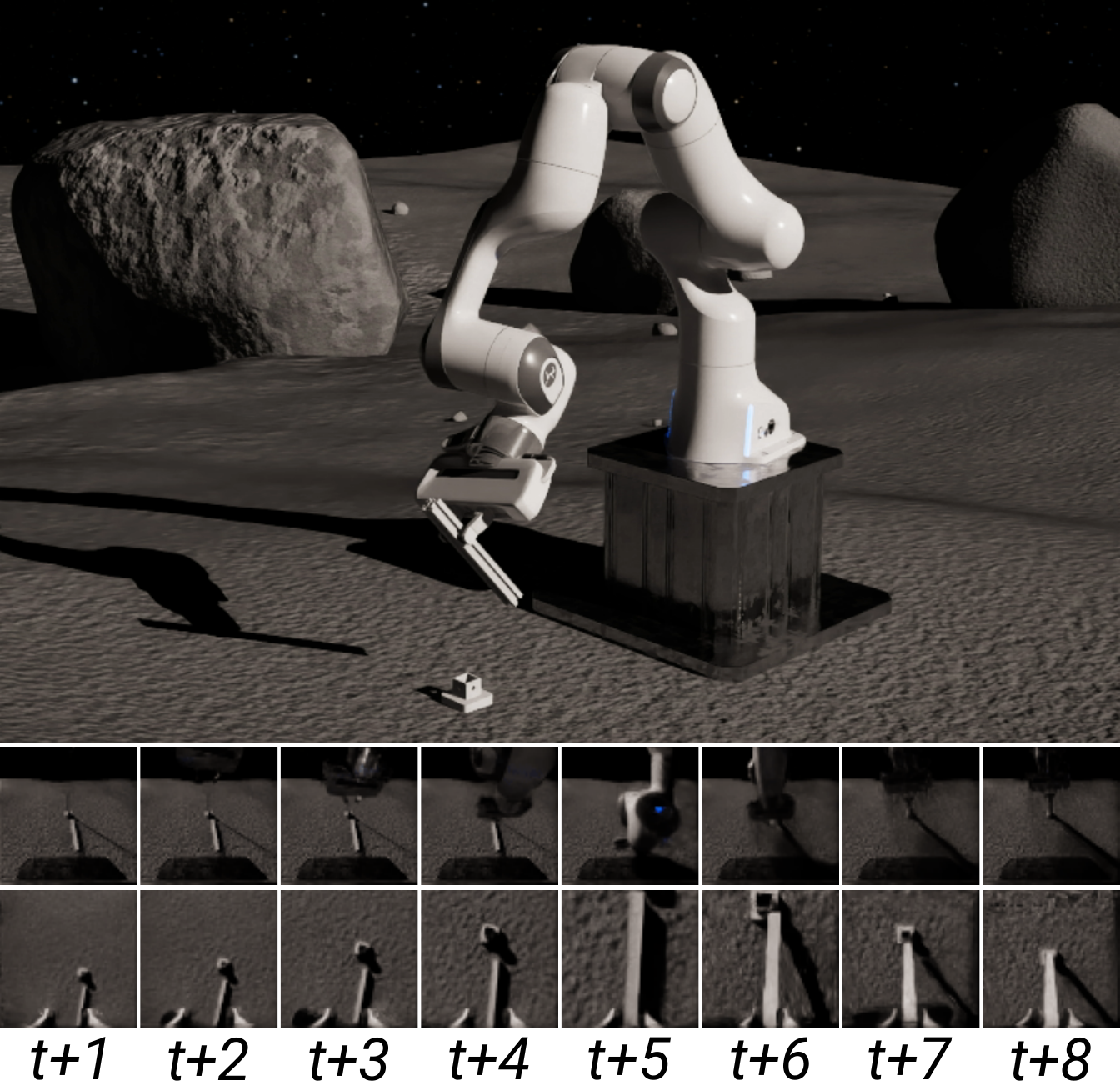}
    }
    \caption{SRB supports training of policies for visuomotor tasks.}
    \label{fig:visuomotor_tasks}
\end{figure}

A key goal for autonomy is to enable agents to act directly from rich sensory inputs, and SRB provides a dedicated testbed for investigating the opportunities of end-to-end visuomotor control. Many complex tasks in the benchmark rely on vision as an essential source of information. As illustrated in Fig.~\ref{fig:visuomotor_tasks}, SRB supports diverse camera configurations tailored to specific challenges. For the \texttt{landing} task, a downward-facing camera provides crucial data for terminal guidance and hazard avoidance on procedural terrain. For contact-rich manipulation like the \texttt{peg\_in\_hole} assembly task, the framework facilitates multi-camera setups that provide both global context from a static base camera and local precision from a wrist-mounted camera. This enables an agent to learn the full pick-and-place sequence, from locating the peg to precisely aligning it for insertion.

\begin{figure}[ht]
    \centering
    \includegraphics[width=\linewidth]{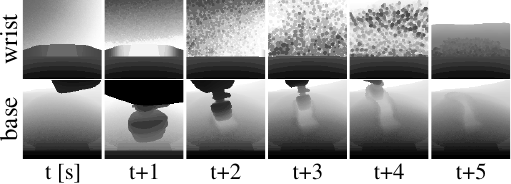}
    \caption{Visual feedback for the \texttt{excavation} scenario, showing depth streams from the base and wrist-mounted cameras over the course of a single episode.}
    \label{fig:excavation_visuals}
\end{figure}

The benefit of vision is particularly evident in scenarios involving complex physical interaction. In the \texttt{excavation} task, augmenting an agent with visual feedback from two simulated depth cameras, as shown in Fig.~\ref{fig:excavation_visuals}, proved beneficial. The rich perceptual data allowed the agent to better coordinate the interaction between its tool and the granular media, resulting in a 93\% increase in the volume of scooped regolith compared to its proprioceptive-only counterpart. These capabilities establish SRB as a versatile platform for developing and benchmarking end-to-end policies across a wide spectrum of perception-driven robotics problems.

\section{Discussion}
\label{sec:discussion}

The experimental results provide a validation of our central hypothesis that robust, generalizable autonomy for space robotics can be achieved through mastering diversity. The successful zero-shot sim-to-real transfer of the rover navigation policy is a confirmation of this procedural paradigm. The physical world is not a perfect replica of any single simulation instance, but rather it is one more unseen variation within a vast distribution of scenarios that the agent is tasked with mastering. This approach reframes the challenge of developing autonomous systems for unknown environments. It moves beyond the limitations of single-point test cases toward a workflow for building statistical assurance. Each procedural scenario acts as a unique hypothesis about the physical world.

This validation clarifies the unique and unmet need that the Space Robotics Bench fills within the broader ecosystem of simulation tools. SRB is not intended to replace the specialized, high-fidelity simulators used for the final V\&V of mission-specific hardware. Instead, it serves as a complementary, open-source platform for the research community. Its purpose is to enable the early-stage, data-intensive development of generalizable policies that are robust by design. By providing a standardized testbed with a focus on diversity, accessibility, and massive parallelism, SRB lowers the barrier to entry for robot learning research in space. Furthermore, as demonstrated by the adaptive control and vision-based case studies, SRB serves as a versatile sandbox for investigating the fundamental interplay between hardware design, control strategies, and sensory modalities.

\subsection*{Limitations}

\begin{figure}[ht]
    \centering
    \subcaptionbox{Simulated Depth Map\label{fig:depth_sim}}{%
        \includegraphics[width=0.48\linewidth]{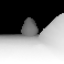}
    }
    \hfill
    \subcaptionbox{Real-World Depth Map\label{fig:depth_real}}{%
        \includegraphics[width=0.48\linewidth]{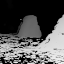}
    }
    \caption{The sim-to-real gap in depth perception, illustrated by the contrast between a clean simulated map and noisy data from a physical camera.}
    \label{fig:depth_gap}
\end{figure}

Despite its demonstrated effectiveness, SRB and the procedural paradigm have limitations that frame important directions for future work. The primary limitation remains the sim-to-real gap, particularly for complex sensory modalities. As shown in Fig.~\ref{fig:depth_gap}, the perceptual gap for vision-based control is a significant hurdle that requires further research into high-fidelity sensor modeling and robust perception algorithms.

Furthermore, our procedural assets, while diverse, should be considered plausible approximations rather than scientifically exact representations of extraterrestrial environments. The goal is to capture the statistical complexity of the real world, not to replicate any specific location perfectly. Similarly, the physics simulation, while powerful, does not model all peculiarities of space environments, such as vacuum conditions, extreme thermal cycles, or radiation effects. These limitations are precisely what motivate our diversity-driven approach. By training policies to be robust to a wide range of unmodeled dynamics and environmental variations, we aim to create systems that can better handle the inevitable discrepancies between any simulation and the true complexities of the final frontier.

\section{Conclusion}
\label{sec:conclusion}

This paper introduced the Space Robotics Bench as an open-source simulation framework designed to enable a new paradigm of developing robust autonomy in space. We argued that for the unknown and unstructured environments of future missions, the most effective path to generalization is not the pursuit of perfect fidelity, but the mastery of immense procedural diversity.

We presented SRB as the tool that realizes this paradigm, providing a modular architecture, a suite of mission-relevant benchmark tasks, and an integrated workflow for generating a virtually unlimited number of unique scenarios. The central contribution of this work is the definitive, empirical validation of this approach. Through the sim-to-real experiment, we demonstrated that a policy trained entirely within diverse procedural environments could be deployed zero-shot to a physical robot. This result provides a validated pathway for creating reliable, learning-based systems for space. By providing this open and accessible platform, we aim to accelerate research, foster collaboration, and help build the trusted autonomous systems that will be essential for our multiplanetary future.

\appendix

\subsection{Hyperparameters}
\label{app:hyperparameters}

Table~\ref{tab:hyperparameters} details the primary hyperparameters used for training all agents. These settings were kept consistent across all tasks to provide a fair and standardized comparison. Any parameters not listed were set to their default values as defined in their respective library implementations.

\newpage

\begin{table}[ht]
    \centering
    \caption{\textsc{Key Hyperparameters of RL Agents.}}
    \label{tab:hyperparameters}
    \addtolength{\tabcolsep}{-0.175em}
    \begin{tabular}{@{}ll@{}}
        \toprule
        \multicolumn{2}{@{}l@{}}{\textbf{PPO}} \vspace{-0.5em}                                         \\
        \quad Learning Rate (Actor \& Critic)       & \(0.0001 \xrightarrow[]{linear} 0.0\) (schedule) \\
        \quad Discount Factor ($\gamma$)            & \(0.997\)                                        \\
        \quad Rollout Buffer Size (per env)         & \(128\)                                          \\
        \quad Minibatch Size                        & \(1024\)                                         \\
        \quad PPO Epochs per Rollout                & \(16\)                                           \\
        \quad GAE Lambda ($\lambda$)                & \(0.95\)                                         \\
        \quad PPO Clip Range ($\epsilon$)           & \(0.2\)                                          \\
        \quad Entropy Coefficient                   & \(0.01\)                                         \\
        \quad Gradient Clipping Norm                & \(0.5\)                                          \\
        \quad Actor/Critic Network Size (MLP Units) & \([384, 384]\)                                   \\
        \midrule
        \multicolumn{2}{@{}l@{}}{\textbf{PPO (LSTM)}}                                                  \\
        \quad LSTM Hidden Size                      & \(384\)                                          \\
        \midrule
        \multicolumn{2}{@{}l@{}}{\textbf{TD3}} \vspace{-0.5em}                                         \\
        \quad Learning Rate (Actor \& Critic)       & \(0.003 \xrightarrow[]{linear} 0.0\) (schedule)  \\
        \quad Discount Factor ($\gamma$)            & \(0.997\)                                        \\
        \quad Replay Buffer Size                    & \(2,000,000\)                                    \\
        \quad Minibatch Size                        & \(512\)                                          \\
        \quad Updates per Environment Step          & \(4\)                                            \\
        \quad Exploration Noise                     & \(\mathcal{N}(0.0, 0.1)\)                        \\
        \quad Target Network Update Rate ($\tau$)   & \(0.005\)                                        \\
        \quad Actor/Critic Network Size (MLP Units) & \([384, 384]\)                                   \\
        \midrule
        \multicolumn{2}{@{}l@{}}{\textbf{DreamerV3}}                                                   \\
        \quad Discount Factor ($\gamma$)            & \(0.997\)                                        \\
        \quad Replay Buffer Size                    & \(2,000,000\)                                    \\
        \quad Batch Size                            & \(16\)                                           \\
        \quad Sequence Length                       & \(64\)                                           \\
        \quad Updates per Environment Step          & \(32\)                                           \\
        \quad Model Size:                           &                                                  \\
        \qquad RSSM Hidden Size                     & \(384\)                                          \\
        \qquad RSSM Deterministic Units             & \(3072\)                                         \\
        \qquad Discrete Latents per State           & \(24\)                                           \\
        \qquad MLP Units                            & \(384\)                                          \\
        \qquad CNN Depth                            & \(24\)                                           \\
        \bottomrule
    \end{tabular}
\end{table}

\balance
\bibliographystyle{style/IEEEtran.bst}
\bibliography{bibliography.bib}

\begin{thebibliography}{10}
\providecommand{\url}[1]{#1}
\csname url@rmstyle\endcsname
\providecommand{\newblock}{\relax}
\providecommand{\bibinfo}[2]{#2}
\providecommand\BIBentrySTDinterwordspacing{\spaceskip=0pt\relax}
\providecommand\BIBentryALTinterwordstretchfactor{4}
\providecommand\BIBentryALTinterwordspacing{\spaceskip=\fontdimen2\font plus
\BIBentryALTinterwordstretchfactor\fontdimen3\font minus
  \fontdimen4\font\relax}
\providecommand\BIBforeignlanguage[2]{{%
\expandafter\ifx\csname l@#1\endcsname\relax
\typeout{** WARNING: IEEEtran.bst: No hyphenation pattern has been}%
\typeout{** loaded for the language `#1'. Using the pattern for}%
\typeout{** the default language instead.}%
\else
\language=\csname l@#1\endcsname
\fi
#2}}

\bibitem{jpl2024enabling}
V.~Verma, J.~Nash, L.~Saldyt, Q.~Dwight, H.~Wang, S.~Myint, J.~Biesiadecki,
  M.~Maimone, A.~Tumbar, A.~Ansar, G.~Kubiak, and R.~Hogg, ``{Enabling Long \&
  Precise Drives for The Perseverance Mars Rover via Onboard Global
  Localization},'' in \emph{{IEEE Aerospace Conference}}, 2024, pp. 1--18.

\bibitem{rognant2019autonomous}
M.~Rognant, C.~Cumer, J.-M. Biannic, M.~A. Roa, A.~Verhaeghe, and
  V.~Bissonnette, ``{Autonomous Assembly of Large Structures in Space: A
  Technology Review},'' in \emph{{European Conference for Aeronautics and
  Aerospace Sciences}}, 2019.

\bibitem{zhihui2021review}
Z.~Xue, J.~Liu, C.~Wu, and Y.~Tong, ``{Review of In-Space Assembly
  Technologies},'' \emph{Chinese Journal of Aeronautics}, vol.~34, no.~11, pp.
  21--47, 2021.

\bibitem{sutton2018reinforcement}
R.~S. Sutton and A.~G. Barto, \emph{Reinforcement {Learning}: {An}
  {Introduction}}.\hskip 1em plus 0.5em minus 0.4em\relax A Bradford Book,
  2018.

\bibitem{gu2025humanoid}
Z.~Gu, J.~Li, W.~Shen, W.~Yu, Z.~Xie, S.~McCrory, X.~Cheng, A.~Shamsah,
  R.~Griffin, C.~K. Liu, A.~Kheddar, X.~B. Peng, Y.~Zhu, G.~Shi, Q.~Nguyen,
  G.~Cheng, H.~Gao, and Y.~Zhao, ``{Humanoid Locomotion and Manipulation:
  Current Progress and Challenges in Control, Planning, and Learning},''
  \emph{arXiv:2501.02116}, 2025.

\bibitem{kim2024openvla}
M.~J. Kim, K.~Pertsch, S.~Karamcheti, T.~Xiao, A.~Balakrishna, S.~Nair,
  R.~Rafailov, E.~P. Foster, P.~R. Sanketi, Q.~Vuong, T.~Kollar, B.~Burchfiel,
  R.~Tedrake, D.~Sadigh, S.~Levine, P.~Liang, and C.~Finn, ``{OpenVLA: An
  Open-Source Vision-Language-Action Model},'' in \emph{Conference on Robot
  Learning}, ser. Proceedings of Machine Learning Research, vol. 270, 2024, pp.
  2679--2713.

\bibitem{cobbe2020leveraging}
K.~Cobbe, C.~Hesse, J.~Hilton, and J.~Schulman, ``{Leveraging Procedural
  Generation to Benchmark Reinforcement Learning},'' in \emph{International
  Conference on Machine Learning}, 2020, pp. 2048--2056.

\bibitem{tobin2017domain}
J.~Tobin, R.~Fong, A.~Ray, J.~Schneider, W.~Zaremba, and P.~Abbeel, ``{Domain
  Randomization for Transferring Deep Neural Networks from Simulation to the
  Real World},'' in \emph{IEEE/RSJ International Conference on Intelligent
  Robots and Systems}, 2017, pp. 23--30.

\bibitem{hughes2014verification}
S.~P. Hughes, R.~H. Qureshi, S.~D. Cooley, and J.~J. Parker, ``{Verification
  and Validation of the General Mission Analysis Tool (GMAT)},'' in
  \emph{AIAA/AAS Astrodynamics Specialist Conference}, 2014.

\bibitem{kenneally2020basilisk}
P.~W. Kenneally, S.~Piggott, and H.~Schaub, ``{Basilisk: A Flexible, Scalable
  and Modular Astrodynamics Simulation Framework},'' \emph{Journal of aerospace
  information systems}, vol.~17, no.~9, pp. 496--507, 2020.

\bibitem{stephenson2024bskrl}
M.~A. Stephenson and H.~Schaub, ``{{BSK-RL}}: {{Modular}}, {{High-Fidelity
  Reinforcement Learning Environments}} for {{Spacecraft Tasking}},'' in
  \emph{75th {{International Astronautical Congress}}}, 2024.

\bibitem{stukes2021innovative}
S.~Stukes, M.~Allan, G.~Bajjalieh, M.~Deans, T.~Fong, J.~Hihn, and H.~Utz, ``An
  innovative approach to modeling viper rover software life cycle cost,'' in
  \emph{IEEE Aerospace Conference}, 2021, pp. 1--16.

\bibitem{largescale2024heli}
C.~Leake, H.~Grip, V.~Steyert, T.~D. Hasseler, M.~Cacan, and A.~Jain,
  ``Helicat-darts: A high fidelity, closed-loop rotorcraft simulator for
  planetary exploration,'' \emph{Aerospace}, vol.~11, no.~9, 2024.

\bibitem{hardeers2024eelsdarts}
T.~D. Hasseler, C.~Leake, A.~Gaut, A.~Elmquist, R.~M. Swan, R.~Royce, B.~Jones,
  B.~Hockman, M.~Paton, G.~Daddi, M.~Ono, R.~Thakker, and A.~Jain,
  ``{EELS-DARTS: A Planetary Snake Robot Simulator for Closed-Loop Autonomy
  Development},'' \emph{Aerospace}, vol.~11, no.~10, 2024.

\bibitem{coltin2019astrobee}
B.~Coltin, A.~Allan, T.~Baron, M.~Bualat, R.~Bundalian, J.~Calero, C.~Colbert,
  M.~D'Souza, S.~J. Deng, C.~Fries, \emph{et~al.}, ``{Astrobee Robot Software:
  A Modern Software System for Space},'' in \emph{IEEE International Conference
  on Robotics and Automation}, 2019, pp. 5556--5562.

\bibitem{Hirano2025IntBall2}
D.~Hirano, S.~Mitani, K.~Watanabe, T.~Nishishita, T.~Yamamoto, and S.~P.
  Yamaguchi, ``{Int-Ball2: On-Orbit Demonstration of Autonomous Intravehicular
  Flight and Docking for Image Capturing and Recharging},'' in \emph{IEEE
  International Conference on Robotics and Automation}, 2025.

\bibitem{antoine2024omnilrs}
A.~Richard, J.~Kamohara, K.~Uno, S.~Santra, D.~van~der Meer,
  M.~Olivares-Mendez, and K.~Yoshida, ``{OmniLRS: A Photorealistic Simulator
  for Lunar Robotics},'' in \emph{IEEE International Conference on Robotics and
  Automation}, 2024, pp. 16\,901--16\,907.

\bibitem{mortensen2024rlroverlab}
A.~B. Mortensen and S.~Bøgh, ``{RLRoverLAB: An Advanced Reinforcement Learning
  Suite for Planetary Rover Simulation and Training},'' in \emph{International
  Conference on Space Robotics}, 2024, pp. 273--277.

\bibitem{el2023rans}
M.~El-Hariry, A.~Richard, and M.~Olivares-Mendez, ``{RANS: Highly-Parallelised
  Simulator for Reinforcement Learning based Autonomous Navigating
  Spacecrafts},'' \emph{arXiv:2310.07393}, 2023.

\bibitem{wang2022collision}
S.~Wang, Y.~Cao, X.~Zheng, and T.~Zhang, ``{Collision-Free Trajectory Planning
  for a 6-DoF Free-Floating Space Robot via Hierarchical Decoupling
  Optimization},'' \emph{IEEE Robotics and Automation Letters}, vol.~7, no.~2,
  pp. 4953--4960, 2022.

\bibitem{orsula2022learning}
A.~Orsula, S.~B{\o}gh, M.~Olivares-Mendez, and C.~Martinez, ``{Learning to
  Grasp on the Moon from 3D Octree Observations with Deep Reinforcement
  Learning},'' in \emph{IEEE/RSJ International Conference on Intelligent Robots
  and Systems}, 2022, pp. 4112--4119.

\bibitem{james2020rlbench}
S.~James, Z.~Ma, D.~R. Arrojo, and A.~J. Davison, ``{RLBench: The Robot
  Learning Benchmark \& Learning Environment},'' \emph{IEEE Robotics and
  Automation Letters}, vol.~5, no.~2, pp. 3019--3026, 2020.

\bibitem{kumar2023robohive}
V.~Kumar, R.~Shah, G.~Zhou, V.~Moens, V.~Caggiano, A.~Gupta, and A.~Rajeswaran,
  ``{RoboHive: A Unified Framework for Robot Learning},'' \emph{Advances in
  Neural Information Processing Systems}, vol.~36, pp. 44\,323--44\,340, 2023.

\bibitem{zhu2020robosuite}
Y.~Zhu, J.~Wong, A.~Mandlekar, R.~Mart\'{i}n-Mart\'{i}n, A.~Joshi, K.~Lin,
  S.~Nasiriany, and Y.~Zhu, ``{robosuite: A Modular Simulation Framework and
  Benchmark for Robot Learning},'' in \emph{arXiv:2009.12293}, 2020.

\bibitem{heo2023furniturebench}
M.~Heo, Y.~Lee, D.~Lee, and J.~J. Lim, ``{FurnitureBench: Reproducible
  Real-World Benchmark for Long-Horizon Complex Manipulation},'' in
  \emph{Robotics: Science and Systems}, 2023.

\bibitem{zakka2023robopianist}
K.~Zakka, P.~Wu, L.~Smith, N.~Gileadi, T.~Howell, X.~B. Peng, S.~Singh,
  Y.~Tassa, P.~Florence, A.~Zeng, and P.~Abbeel, ``{RoboPianist: Dexterous
  Piano Playing with Deep Reinforcement Learning},'' in \emph{Conference on
  Robot Learning}, ser. Proceedings of Machine Learning Research, vol. 229,
  2023, pp. 2975--2994.

\bibitem{sferrazza2024humanoidbench}
C.~Sferrazza, D.-M. Huang, X.~Lin, Y.~Lee, and P.~Abbeel, ``{HumanoidBench:
  Simulated Humanoid Benchmark for Whole-Body Locomotion and Manipulation},''
  \emph{arXiv:2403.10506}, 2024.

\bibitem{koutras2021marsexplorer}
D.~I. Koutras, A.~C. Kapoutsis, A.~A. Amanatiadis, and E.~B. Kosmatopoulos,
  ``\BIBforeignlanguage{en}{{MarsExplorer}: {Exploration} of {Unknown}
  {Terrains} via {Deep} {Reinforcement} {Learning} and {Procedurally}
  {Generated} {Environments}},'' \emph{\BIBforeignlanguage{en}{Electronics}},
  vol.~10, no.~22, p. 2751, 2021.

\bibitem{han2024fetchbench}
B.~Han, M.~Parakh, D.~Geng, J.~A. Defay, G.~Luyang, and J.~Deng, ``{FetchBench:
  A Simulation Benchmark for Robot Fetching},'' in \emph{Conference on Robot
  Learning}, ser. Proceedings of Machine Learning Research, vol. 270, 2024, pp.
  3053--3071.

\bibitem{mittal2023orbit}
M.~Mittal, C.~Yu, Q.~Yu, J.~Liu, N.~Rudin, D.~Hoeller, J.~L. Yuan, R.~Singh,
  Y.~Guo, H.~Mazhar, A.~Mandlekar, B.~Babich, G.~State, M.~Hutter, and A.~Garg,
  ``{Orbit: A Unified Simulation Framework for Interactive Robot Learning
  Environments},'' \emph{IEEE Robotics and Automation Letters}, vol.~8, no.~6,
  pp. 3740--3747, 2023.

\bibitem{macenski2022ros2}
S.~Macenski, T.~Foote, B.~Gerkey, C.~Lalancette, and W.~Woodall, ``Robot
  {Operating} {System} 2: {Design}, architecture, and uses in the wild,''
  \emph{Science Robotics}, vol.~7, no.~66, 2022.

\bibitem{towers2024gymnasium}
M.~Towers, A.~Kwiatkowski, J.~Terry, J.~U. Balis, G.~De~Cola, T.~Deleu,
  M.~Goul{\~a}o, A.~Kallinteris, M.~Krimmel, A.~KG, \emph{et~al.},
  ``{Gymnasium: A Standard Interface for Reinforcement Learning
  Environments},'' \emph{arXiv:2407.17032}, 2024.

\bibitem{raffin2021stable}
A.~Raffin, A.~Hill, A.~Gleave, A.~Kanervisto, M.~Ernestus, and N.~Dormann,
  ``{Stable-Baselines3: Reliable Reinforcement Learning Implementations},''
  \emph{Journal of Machine Learning Research}, vol.~22, no. 268, pp. 1--8,
  2021.

\bibitem{serrano2023skrl}
A.~Serrano-Mu\~{n}oz, D.~Chrysostomou, S.~B{\o}gh, and N.~Arana-Arexolaleiba,
  ``{skrl: Modular and Flexible Library for Reinforcement Learning},''
  \emph{Journal of Machine Learning Research}, vol.~24, no.~1, 2023.

\bibitem{pyo3}
\BIBentryALTinterwordspacing
{PyO3 Project and Contributors}, ``{PyO3: Rust bindings for the Python
  interpreter},'' 2025. [Online]. Available: \url{https://github.com/PyO3/pyo3}
\BIBentrySTDinterwordspacing

\bibitem{ansel2024pytorch}
J.~Ansel, E.~Yang, H.~He, N.~Gimelshein, A.~Jain, M.~Voznesensky, B.~Bao,
  P.~Bell, D.~Berard, E.~Burovski, G.~Chauhan, A.~Chourdia, W.~Constable,
  A.~Desmaison, Z.~DeVito, E.~Ellison, W.~Feng, J.~Gong, M.~Gschwind, B.~Hirsh,
  S.~Huang, K.~Kalambarkar, L.~Kirsch, M.~Lazos, M.~Lezcano, Y.~Liang,
  J.~Liang, Y.~Lu, C.~Luk, B.~Maher, Y.~Pan, C.~Puhrsch, M.~Reso, M.~Saroufim,
  M.~Y. Siraichi, H.~Suk, M.~Suo, P.~Tillet, E.~Wang, X.~Wang, W.~Wen,
  S.~Zhang, X.~Zhao, K.~Zhou, R.~Zou, A.~Mathews, G.~Chanan, P.~Wu, and
  S.~Chintala, ``{PyTorch 2: Faster Machine Learning Through Dynamic Python
  Bytecode Transformation and Graph Compilation},'' in \emph{29th ACM
  International Conference on Architectural Support for Programming Languages
  and Operating Systems, Volume 2}, 2024.

\bibitem{blender}
\BIBentryALTinterwordspacing
{Blender Foundation and Community}, ``{Blender},'' 2025. [Online]. Available:
  \url{http://blender.org}
\BIBentrySTDinterwordspacing

\bibitem{zhou2019continuity}
Y.~Zhou, C.~Barnes, J.~Lu, J.~Yang, and H.~Li, ``{On the Continuity of Rotation
  Representations in Neural Networks},'' in \emph{IEEE/CVF Conference on
  Computer Vision and Pattern Recognition}, 2019.

\bibitem{schulman2017proximal}
J.~Schulman, F.~Wolski, P.~Dhariwal, A.~Radford, and O.~Klimov, ``{Proximal
  Policy Optimization Algorithms},'' \emph{arXiv:1707.06347}, 2017.

\bibitem{fujimoto2018addressing}
S.~Fujimoto, H.~Hoof, and D.~Meger, ``{Addressing Function Approximation Error
  in Actor-Critic Methods},'' in \emph{International Conference on Machine
  Learning}, 2018, pp. 1587--1596.

\bibitem{hafner2025mastering}
D.~Hafner, J.~Pasukonis, J.~Ba, and T.~Lillicrap, ``{Mastering Diverse Control
  Tasks through World Models},'' \emph{Nature}, vol. 640, pp. 647--653, 2025.

\bibitem{khatib1987osc}
O.~Khatib, ``{A Unified Approach for Motion and Force Control of Robot
  Manipulators: The Operational Space Formulation},'' \emph{IEEE Journal on
  Robotics and Automation}, vol.~3, no.~1, pp. 43--53, 1987.

\end{thebibliography}
\vfill
\end{document}